\documentclass[preprint, review, 10pt]{elsarticle}

\usepackage{ifluatex}
\ifluatex
\usepackage[utf8]{luainputenc}
\usepackage[OT1]{fontenc}
\else
\usepackage[T1]{fontenc}
\fi

\usepackage{graphicx}
\graphicspath{{./figs/}}
\DeclareGraphicsExtensions{.png,.pdf}
\usepackage{xcolor}
\usepackage[cmex10]{amsmath}
\usepackage{amsthm,amssymb,amsfonts,mathrsfs,bm,mathtools,cases}
\usepackage{subfig}
\usepackage{etoolbox}
\usepackage[linesnumbered,vlined,ruled,commentsnumbered]{algorithm2e}
\usepackage{setspace}
\usepackage{xspace}
\usepackage[inline]{enumitem}

\makeatletter
\newcommand*{\ie}{%
  \@ifnextchar{,}%
  {{i.e.}}%
  {{i.e.,}\@\xspace}%
}
\newcommand*{\eg}{%
  \@ifnextchar{,}%
  {{e.g.}}%
  {{e.g.,}\@\xspace}%
}
\newcommand*{\etc}{%
  \@ifnextchar{.}%
  {{etc}}%
  {{etc.}\@\xspace}%
}
\newcommand*{\etal}{%
  \@ifnextchar{.}%
  {{et al}}%
  {{et al.}\@\xspace}%
}
\newcommand*{\cf}{%
  \@ifnextchar{.}%
  {{cf}}%
  {{cf.}\@\xspace}%
}
\newcommand*{\aka}{%
  \@ifnextchar{,}%
  {{a.k.a.}}%
  {{a.k.a.}\@\xspace}%
}
\makeatother

\usepackage{pgfplots}
\usepackage{pgfplotstable}
\usepackage{booktabs}
\usepackage{threeparttable}
\usepackage{multirow}
\usepackage{multicol}
\pgfplotsset{compat = newest}
\colorlet{negro}{black}
\colorlet{gris}{black!70}
\colorlet{rojo}{red!70!black}
\colorlet{rojol}{red}
\usepackage{tabu}
\usetikzlibrary{
  calc,
  trees,
  shadows,
  positioning,
  arrows,
  arrows.meta,
  chains,
  decorations.pathreplacing,
  decorations.pathmorphing,
  decorations.shapes,
  decorations.text,
  decorations.markings,
  shapes,
  shapes.geometric,
  shapes.symbols,
  matrix,
  patterns,
  intersections,
  fit
}

\newcommand{\Integer}{\mathbb{Z}}
\newcommand{\IntegerP}{\mathbb{Z}_{\geq 0}}
\newcommand{\IntegerPP}{\mathbb{Z}_{>0}}
\newcommand{\Real}{\mathbb{R}}

\newcommand{\RealPP}{\mathbb{R}_{>0}}

\newcommand\given{{\mathbin{}\mid\mathbin{}}}
\newcommand\vect[1]{\mathbf{#1}}

\providecommand\given{} 
\newcommand\SetSymbol[1][]{
  \nonscript\,#1\vert \allowbreak \nonscript\,\mathopen{}}
\DeclarePairedDelimiterX\Set[1]{\lbrace}{\rbrace}%
{ \renewcommand\given{\SetSymbol[\delimsize]} #1 }
\DeclarePairedDelimiterX\innerp[2]{\langle}{\rangle}{#1
  \mathop{}\delimsize\vert\mathop{} #2}
\DeclarePairedDelimiterX\norm[1]\lVert\rVert{\ifblank{#1}{\:\cdot\:}{#1}}

\DeclareMathOperator{\Expect}{\mathbb{E}}
\DeclareMathOperator{\Prob}{\mathbb{P}}

\DeclareMathOperator{\rank}{rank}

\DeclareMathOperator*{\Argmin}{arg\,min}

\DeclareMathOperator{\Kprod}{\otimes_{\mathcal{H}}}

\theoremstyle{definition}
\newtheorem{prop}{Proposition}

\newcommand{\Rmnum}[1]{\expandafter\@slowromancap\romannumeral #1@}

\makeatletter
\newcommand*{\addFileDependency}[1]{
  \typeout{(#1)}
  \@addtofilelist{#1}
  \IfFileExists{#1}{}{\typeout{No file #1.}}
}
\makeatother

\renewenvironment{thebibliography}[1]{
  \begin{oldthebibliography}{#1}
    \setstretch{1.3}
    \setlength{\itemsep}{.1em}
    \setlength{\parskip}{0em}
  }
  {
  \end{oldthebibliography}
}

\usepackage{lineno}

\newcommand{\proofofref}{}
\newproof{zproofof}{Proof of \proofofref}
\newenvironment{proofof}[1]
 {\renewcommand{\proofofref}{#1}\zproofof}
 {\endzproofof}

\usepackage{xr}
\allowdisplaybreaks
\biboptions{sort&compress}

\journal{Signal Processing}
\usepackage{natbib}
\begin{document}

\begin{frontmatter}


  \title{Clustering in Networks Via\\ Kernel-ARMA Modeling and the
    Grassmannian:\\ The Brain-Network Case}



  \author[label1]{Cong Ye\corref{cor1}}
  \address[label1]{Department of Electrical Engineering, University at Buffalo,
    The State University of New York (SUNY), NY 14260, USA.} 
    \address[label2]{Neuroscience Program, University at Buffalo, Buffalo, NY 14260,USA}
  \address[label3]{Department of Mathematics and the Computational and
    Data-Enabled Science and Engineering Program, University at Buffalo, SUNY,
    NY 14260, USA.\fnref{label5}} 

  \address[label4]{Department of Psychology, Drexel University, PA 19104, USA,
    and the Perelman School of Medicine, University of Pennsylvania, PA 19104,
    USA.}

  \cortext[cor1]{Corresponding author}
  \ead{congye@buffalo.edu}
  \author[label1]{Konstantinos~Slavakis}
  \author[label1]{Pratik V.~Patil}
  \author[label2]{Johan~Nakuci}
  \author[label3]{Sarah~F.~Muldoon}
  \author[label4]{John~Medaglia}

  \begin{abstract}

    This paper introduces a clustering framework for networks with nodes are annotated with time-series data. The framework addresses all types of
    network-clustering problems: State clustering, node clustering within states
    (a.k.a.\ topology identification or community detection), and even
    subnetwork-state-sequence identification/tracking. Via a bottom-up approach,
    features are first extracted from the raw nodal time-series data by kernel
    autoregressive-moving-average modeling to reveal non-linear dependencies
    and low-rank representations, and then mapped onto the Grassmann manifold
    (Grassmannian). All clustering tasks are performed by leveraging the
    underlying Riemannian geometry of the Grassmannian in a novel way. To
    validate the proposed framework, brain-network clustering is considered,
    where extensive numerical tests on synthetic and real
    functional Magnetic Resonance Imaging (fMRI) data demonstrate that the
    advocated learning framework compares favorably versus several
    state-of-the-art clustering schemes.
  \end{abstract}


\end{frontmatter}

\section{Introduction}

\subsection{Background}

\begin{figure}[!t]
  \centering
  \includegraphics[width = .65\linewidth]{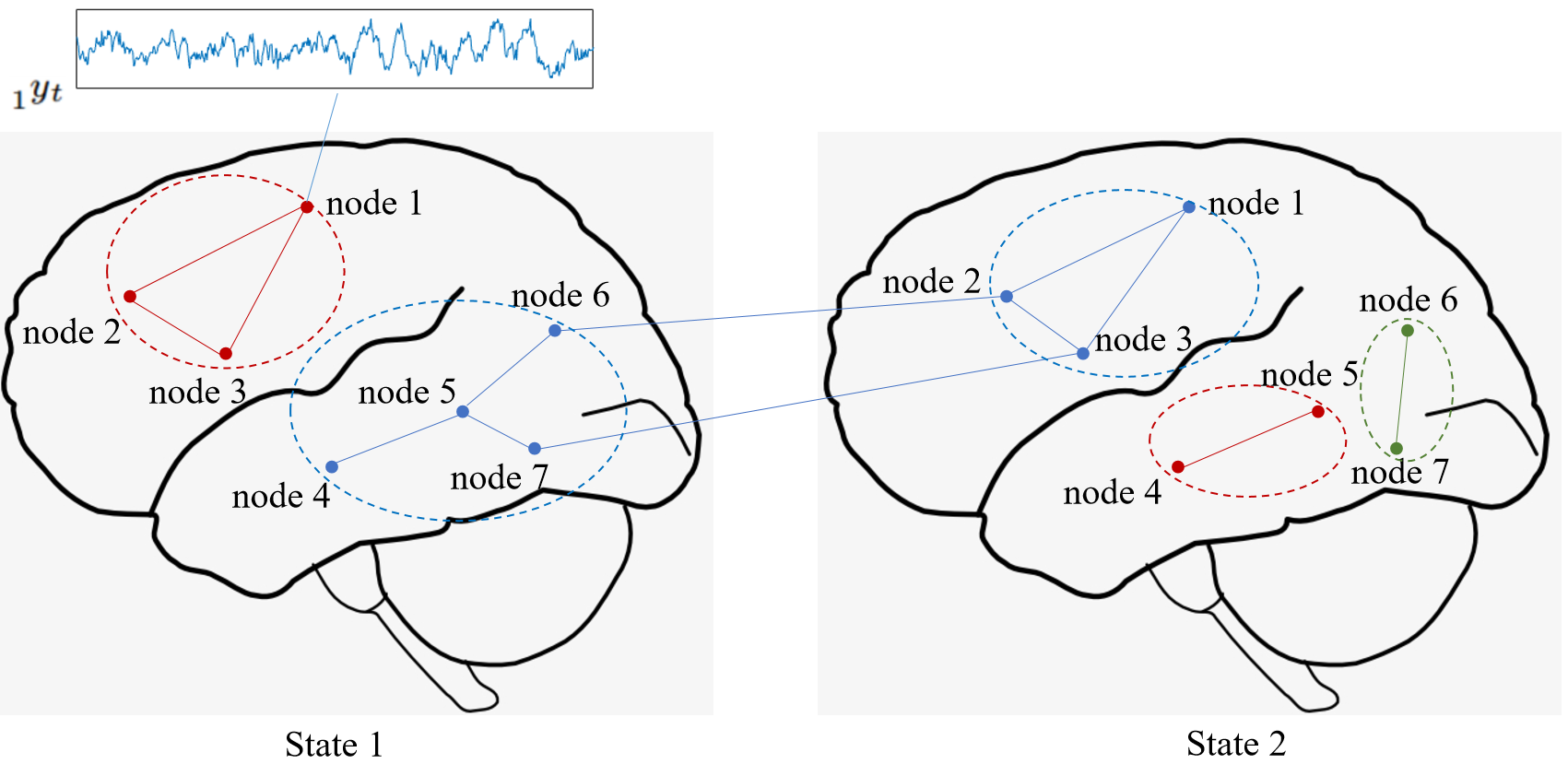}
  \caption{States, communities and subnetwork state sequences in brain
    networks. Nodes connected by soild line are driven by a common latent
    (stochastic) process. The ``blue'' nodes in states 1 and 2 are driven by a common latent
    (stochastic) process and they belong to same sub-network.  }\label{Fig:Track.subnetwork}
\end{figure}

Network clustering is the task of assigning nodes to groups via user-defined
(statistical) ``similarities'' among nodal time series (signals), and is ubiquitous across a plethora of disciplines such as computer
vision~\citep{fan2018unsupervised}, wireless-sensor~\citep{abuarqoub2017dynamic},
social~\citep{huang2016exploring} and brain networks~\citep{martens2017brain}. In brain networks, the choice of scale and type of data determine how networks are built.  At the microscopic level, network nodes might be neurons, and edges could represent anatomical connections such as synapses (structural connectivity), or statistical relationships between firing patterns of neurons (functional connectivity).  Similarly, at the macroscopic level, nodes can represent brain regions. At this scale, in structural networks, edges might represent long range anatomical connections between brain regions or, in functional networks, statistical relationships between regional brain dynamics recorded via functional Magnetic Resonance Imaging (fMRI) or encephalopathy (EEG).  Here, we are interested in functional brain networks in which network nodes represent brain regions whose activity can be represented by a time series describing the dynamic evolution of brain activity.\citep{Feldt:2011fh}; \eg,
Fig.~\ref{Fig:Track.subnetwork}. In the brain-network context, network
clustering has been instrumental in verifying and describing the dynamic nature of brain
networks, as well as in detecting and predicting brain disorders such as
epilepsy~\citep{pedersen2015increased}, schizophrenia~\citep{Broyd.default.09},
Alzheimer disease and autism~\citep{Stam.Alzheimer.09}.

Network clustering aims at three primary goals: State clustering, node
clustering within a given state (a.k.a.\ community detection or topology
identification), and subnetwork-state-sequence clustering/tracking. Loosely
speaking, a ``state'' corresponds to a specific network-wide (``global'')
network topology or nodal connectivity pattern which stays fixed over a time
interval. For example, Fig.~\ref{Fig:Track.subnetwork} depicts two states of a given brain network, with
distinct nodal connectivity patterns. Node clustering
parcellates nodes within a state via ``similarities'' of their time series. Two
communities can be seen in the first state, while three communities emerge in
the second state of Fig.~\ref{Fig:Track.subnetwork}. Furthermore, a ``subnetwork
state sequence'', defined as the latent (stochastic) process that drives a
subnetwork/subgroup of nodal time series, may span several ``global'' states,
and the collaborating nodes may even change as the network topology transitions
from one state to another. For example, it is conceivable that a specific latent
(stochastic) process spans different states of a brain network to drive the
time-series data of the ``blue'' nodes in Fig.~\ref{Fig:Track.subnetwork}.

\subsection{Prior Art}


Most network-clustering methods are used for state and nodal clustering, while
only very few schemes identify/track subnetwork state sequences. To avoid an
exhaustive list of references, only a few examples on state clustering are
mentioned here. Studies~\citep{anderson2014independent, ma2014dynamic} utilize
independent vector analysis and K-means to detect changes in connectivity
patterns. Moreover, \citep{ou2015characterizing, zheng2017student} advocate
hidden Markov models to characterize and cluster network-topology
dynamics/states, while \citep{Masuda:ScientificReports:19} applies hierarchical
clustering onto a time series of graph-distance measures to identify discrete
states of networks.

Node clustering (a.k.a.\ community detection or topology identification) has
been studied extensively for both static and dynamic networks. Modularity
maximization~\citep{lu2015parallel, xiang2016local} is by-now a classical method
for community detection. In~\citep{orhan2012epileptic}, K-means is applied onto
the wavelet coefficients of nodal signals, while \citep{martens2017brain,
  li2017motif} promote network ``motifs'' as features to detect network
communities. In~\citep{mammone2011clustering}, EEG-data topography via Renyi's
entropy was proposed as a feature extraction mapping, before applying
self-organizing maps as the off-the-shelf clustering algorithm. In the recently
popular graph-signal-processing context~\citep{Mateos:SPM:19, Segarra:17},
topology inference is achieved by solving optimization problems formed via the
Laplacian matrix of the network. Moreover, motivated by the observation that
changes in nodal communities suggest changes in network states,
\citep{al2018tensor} uses fMRI data to perform community detection, and
subsequently state clustering, by capitalizing on K-means, multi-layer modeling,
(Tucker) tensor and higher-order singular value decompositions.

There are only few methods that can cluster subnetwork state sequences,
especially in the brain-network context. In~\citep{brechet2019capturing},
features extracted from the frequency content of time series are fed into the
classical K-means to yield the subnetwork state sequences. A computer-vision
approach is introduced in~\citep{lim2018novel} where time series data are
transformed into dynamic topographic maps via motion vectors. 


\subsection{Contributions}

The contributions of this manuscript are as follows:

\begin{enumerate}[label = \textbf{(\roman*)}, wide, labelwidth=!,
  labelindent=0pt]

\item By capitalizing on the directions established
  by~\citep{slavakis2018clustering}, a \textit{unifying}\/ clustering framework
  with strong geometric flavor is introduced that makes no assumptions on the
  network's stationarity and can carry through \textit{all}\/ possible
  brain-clustering duties, \ie, state and node clustering, as well as
  subnetwork-state-sequence tracking.

\item A kernel (vector-valued) autoregressive-moving-average (K-ARMA) model,
  which appears to be novel in the network-science literature, is proposed to
  capture latent non-linear and causal dependencies among network
  time-series. This K-ARMA model propels the network-feature extraction of any
  network-clustering task in this article. Per application of the K-ARMA model,
  a system-identification problem is solved to extract a low-rank observability
  matrix. Features are defined as the low-rank column spaces of those
  observability matrices. For a fixed rank, those features become points of the
  Grassmann manifold (Grassmannian), which enjoys the rich Riemannian geometry.

\item The framework assumes no prior knowledge on affinity/adjacency matrices of
  the network, as it is customary done in the literature; \eg, Laplacian
  matrices~\citep{mateos2019connecting}. All such information can be computed
  \textit{from scratch}\/ in the proposed framework via the K-ARMA
  feature-extraction scheme.

\item Having computed features, the Riemannian multi-manifold modeling
  (RMMM)~\citep{RMMM.AISTATS.15, RMMM.arxiv, slavakis2018clustering} postulates
  that clusters take the form of sub-manifolds in the Grassmannian. To identify
  clusters, the underlying Riemannian geometry is exploited by the
  geodesic-clustering-with-tangent-spaces (GCT) algorithm~\citep{RMMM.AISTATS.15,
    RMMM.arxiv, slavakis2018clustering}. Unlike the standard practice of using
  only the Riemannian distance, \eg, \citep{yger2017riemannian}, GCT considers
  \textit{both} distance and angular information to improve clustering accuracy.

\item In contrast to~\citep{RMMM.AISTATS.15, RMMM.arxiv, slavakis2018clustering},
  where the number of clusters needs to be known a priori, this paper
  incorporates hierarchical clustering to render GCT \textit{free}\/ from any
  a-priori knowledge of the number of clusters.

\item Extensive numerical tests on synthetic and real fMRI data demonstrate that
  the proposed framework compares favorably versus state-of-the-art
  manifold-learning and brain-network clustering schemes.

\end{enumerate}

\begin{figure}[!t]
  \centering
  \includegraphics[width = \linewidth]{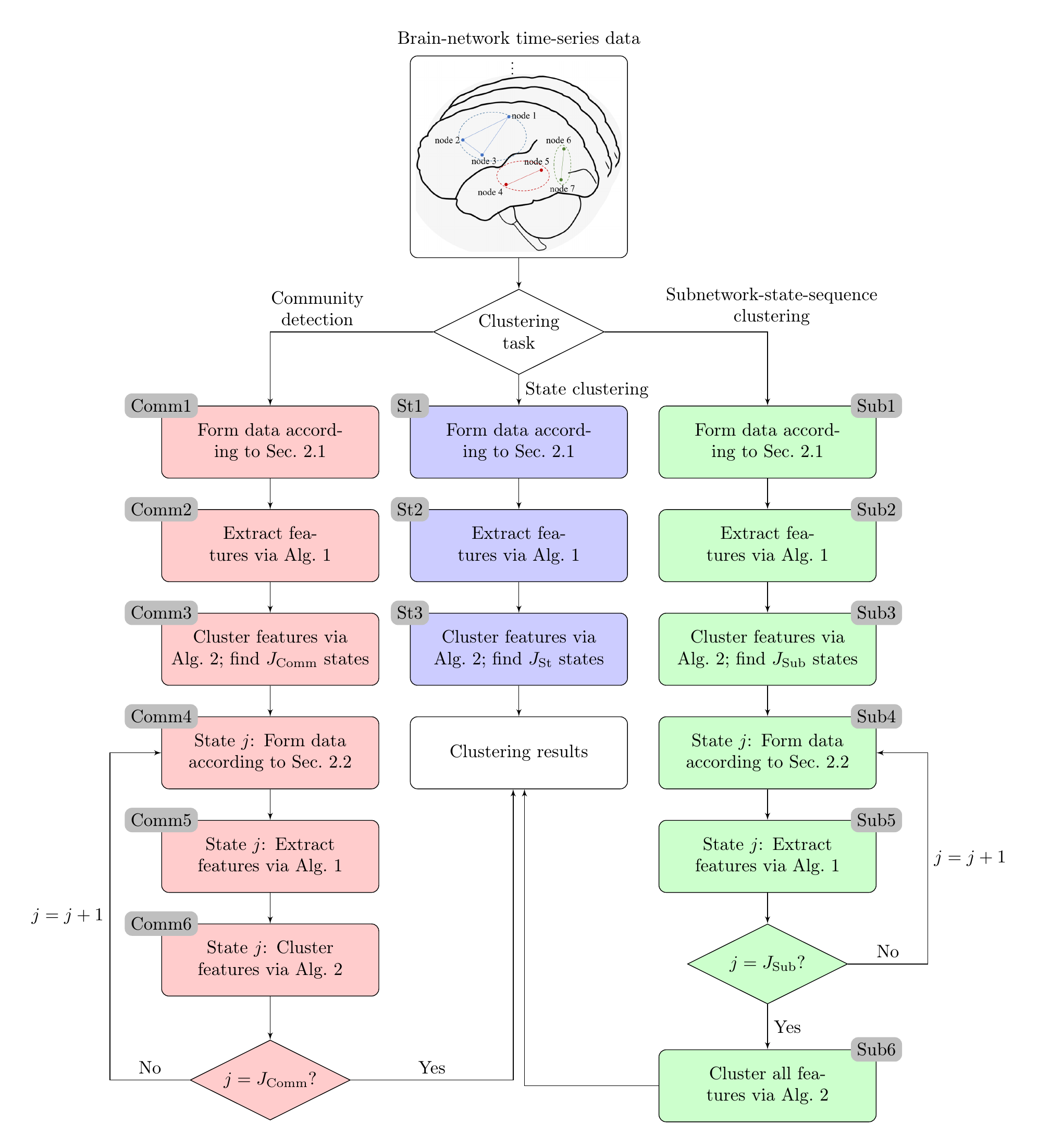}
  \caption{The pipeline of the proposed clustering
    framework.}\label{Fig:Flowchart}
\end{figure}

For convenience, the proposed clustering framework is summarized in
Fig.~\ref{Fig:Flowchart}, and its building blocks, or modules, are delineated in
the rest of the paper. The K-ARMA model and the feature-extraction mechanism are
introduced in Section~\ref{Sec:K-ARMA}. The new variant of the GCT clustering
algorithm is presented in Section~\ref{Sec:Clustering}, while numerical tests on
synthetic and real fMRI data are showed in Section~\ref{Sec:Tests}. Numerical
tests and results that do not fit in the main manuscript are deferred to the
supplementary file. Sections, figures, and tables of the supplementary manuscript
are marked with the ``S'' qualifier.

\section{Network-Feature Extraction by Kernel-ARMA Modeling}\label{Sec:K-ARMA}

Consider a (brain) network/graph
$\mathcal{G} \coloneqq (\mathcal{N}, \mathcal{E})$, with sets of nodes
$\mathcal{N}$, of cardinality $\lvert\mathcal{N} \rvert$, and edges
$\mathcal{E}$. Each node $\nu\in \mathcal{N}$ is annotated with a stochastic
process (time series) $(\prescript{}{\nu}{y}_{t})_{t\in \Integer}$, where $t$
denotes discrete time and $\Integer$ the set of all integer numbers; \cf
Fig.~\ref{Fig:Track.subnetwork}. To avoid congestion in notations,
$\prescript{}{\nu}{y}_{t}$ stands for both the random variable (RV) and its
realization. In fMRI, nodes $\mathcal{N}$ comprise regions of interest (ROI) of
the brain which are created either anatomically or functionally, and
$(\prescript{}{\nu}{y}_{t})_{t\in \Integer}$ becomes a blood-oxygen-level
dependent (BOLD) time series~\citep{ogawa1990brain}, \eg,
Fig.~\ref{Fig:fMRI.BOLD}. For index $\mathcal{V} \subset \mathcal{N}$ and
$q\in\IntegerPP$, the $q\times 1$ vector
$\prescript{}{\mathcal{V}}{\vect{y}}_{t}$ is used in this manuscript to collect
all signal samples from node(s) $\mathcal{V}$ of the network at time $t$, and to
unify several scenarios of interest as the following discussion demonstrates.

\subsection{State Clustering ($\mathcal{V} \coloneqq \mathcal{N}$)}
\label{Sec:state.clustering}

Since a ``state'' is a global attribute of the network, vector
$\prescript{}{\mathcal{N}}{\vect{y}}_{t} \coloneqq [\prescript{}{1}{y}_t,
\ldots, \prescript{}{\lvert \mathcal{N} \rvert}{y}_t]^{\intercal}$, with
$\mathcal{V} \coloneqq \mathcal{N}$ and $q \coloneqq \lvert \mathcal{N} \rvert$,
stands as the "snapshot" of the network at time $t$. The time series
$(\prescript{}{\mathcal{N}}{\vect{y}}_{t})_t$ are the data formed in modules St1,
Comm1 and Sub1 of Fig.~\ref{Fig:Flowchart}. 

\subsection{Community Detection and  Subnetwork-State-Sequence Clustering
  ($\mathcal{V} \coloneqq \nu$)} \label{Sec:comm.task.clustering}

In the case of community detection and subnetwork-state-sequence clustering,
nodes $\mathcal{N}$ need to be partitioned through the (dis)similarities of
their time series. To detect common features and to identify those nodes, it is
desirable first to extract individual features from each nodal time series. To
this end, $\mathcal{V}$ is assigned the value $\nu$, so that
$\forall \nu\in\mathcal{N}$, for a given buffer length
$\texttt{Buff}_{\nu}\in \IntegerPP$ and with $q = \texttt{Buff}_{\nu}$,
$\prescript{}{\nu}{\vect{y}}_{t}$ takes the form of
$[\prescript{}{\nu}{y}_t, \ldots, \prescript{}{\nu}{y}_{t+ \texttt{Buff}_{\nu}
  -1}]^{\intercal}$. If $\mathfrak{T}_j$ comprises all time indices of the $j$th
state of a network, then the time series
$\Set{(\prescript{}{\nu}{\vect{y}}_{t})_{t\in
    \mathfrak{T}_j}}_{\nu\in\mathcal{N}}$ are the data formed in modules Comm4
and Sub4 of Fig.~\ref{Fig:Flowchart}. 

\subsection{Extracting Grassmannian Features}\label{Sec:KARMA.features}

Consider now a user-defined RKHS $\mathcal{H}$ with its kernel mapping
$\varphi(\cdot)$; \cf~Sec.~\ref{App:RKHS}. Given $N\in\IntegerPP$ and assuming
that the sequence $(\prescript{}{\mathcal{V}}{\vect{y}}_{t})_t$ is available,
define
$\bm{\varphi}_t \coloneqq [\varphi(\prescript{}{\mathcal{V}}{\vect{y}}_{t}),
\varphi(\prescript{}{\mathcal{V}}{\vect{y}}_{t+1}), \ldots,
\varphi(\prescript{}{\mathcal{V}}{\vect{y}}_{t+N-1})]^{\intercal} \in
\mathcal{H}^N$. This work proposes the following kernel (K-)ARMA model to fit
the variations of features $\Set{\bm{\varphi}_t}_t$ within space
$\mathcal{H}^N$: There exist matrices $\vect{C}\in \Real^{N\times\rho}$,
$\vect{A}\in \Real^{\rho\times \rho}$, the latent variable
$\bm{\psi}_t\in \mathcal{H}^{\rho}$, and vectors
$\bm{\upsilon}_t\in \mathcal{H}^N$, $\bm{\omega}_t\in \mathcal{H}^{\rho}$ that
capture noise and approximation errors, s.t.\ $\forall t$,
\begin{subequations}\label{KARMA}
  \begin{align}
    \bm{\varphi}_t & = \vect{C} \bm{\psi}_t + \bm{\upsilon}_t \,, \\
    \bm{\psi}_t & = \mathbf{A} \bm{\psi}_{t-1} + \bm{\omega}_t \,.
  \end{align}
\end{subequations}

Kernel-based ARMA models have been already studied in the context of
support-vector regression~\citep{Drezet.98, martinez2006support,
  shpigelman2009kernel}. However, those models are different than \eqref{KARMA}
since only the AR and MA vectors of coefficients are mapped to an RKHS feature
space, while the observed data $\prescript{}{\nu}{y}_{t}$ (of only a single time
series) are kept in the input space. Here, \eqref{KARMA} offers a way to map
even the observed data to an RKHS to capture non-linearities in data via
applying the ARMA idea to properly chosen feature spaces. In a different
context~\citep{Pincombe:ASOR}, time series of graph-distance metrics are fitted
by ARMA modeling to detect anomalies and thus identify states in
networks. Neither the Grassmannian nor kernel functions were investigated
in~\citep{Pincombe:ASOR}.

\begin{prop}\label{Prop:O}
  Given parameter $m\in\IntegerPP$, define the ``forward'' matrix-valued
  function
  \begin{subequations}\label{FB} 
    \begin{align}%
      \bm{\mathcal{F}}_t
      & \coloneqq \begin{bmatrix}
        \bm{\varphi}_t & \bm{\varphi}_{t+1} & \ldots &
        \bm{\varphi}_{t+\tau_{\text{f}}-1} \\ 
        \bm{\varphi}_{t+1} & \bm{\varphi}_{t+2} & \ldots &
        \bm{\varphi}_{t+\tau_{\text{f}}} \\ 
        \vdots & \vdots & \ddots & \vdots \\ 
        \bm{\varphi}_{t+m-1} & \bm{\varphi}_{t+m} & \ldots &
        \bm{\varphi}_{t+\tau_{\text{f}}+m-2} 
      \end{bmatrix} \in \mathcal{H}^{mN \times \tau_{\text{f}}} \,,
    \end{align}
    and the ``backward'' matrix-valued function
    \begin{align}
      \bm{\mathcal{B}}_t
      & \coloneqq \begin{bmatrix}
        \bm{\varphi}_t & \bm{\varphi}_{t+1} & \ldots &
        \bm{\varphi}_{t+\tau_{\text{f}}-1} \\ 
        \bm{\varphi}_{t-1} & \bm{\varphi}_{t} & \ldots &
        \bm{\varphi}_{t+\tau_{\text{f}}-2} \\ 
        \vdots & \vdots & \ddots & \vdots \\ 
        \bm{\varphi}_{t-\tau_{\text{b}}+1} & \bm{\varphi}_{t-\tau_{\text{b}}+2}
        & \ldots & \bm{\varphi}_{t+\tau_{\text{f}}-\tau_{\text{b}}}
      \end{bmatrix} \in\mathcal{H}^{\tau_{\text{b}}N\times \tau_{\text{f}}} \,.
    \end{align}%
  \end{subequations}%
  Then, there exist matrices
  $\bm{\Pi}_{t+1}\in \Real^{\rho\times \tau_{\text{b}}N}$ and
  $\bm{\mathcal{E}}_{t+1}^{\tau_{\text{f}}}\in \Real^{mN\times
    \tau_{\text{b}}N}$ s.t.\ the following \textit{low-rank}\/ factorization
  holds true:
  \begin{align}
    \tfrac{1}{\tau_{\text{f}}} \bm{\mathcal{F}}_{t+1} \Kprod
    \bm{\mathcal{B}}_t^{\intercal} = \vect{O} \bm{\Pi}_{t+1} +
    \bm{\mathcal{E}}_{t+1}^{\tau_{\text{f}}} \,,\label{low.rank.formula}
  \end{align}
  where product $\Kprod$ is defined in Sec.~\ref{App:RKHS}, and $\vect{O}$
  is the so-called 
  \textit{observability}\/ matrix:
  $\vect{O} \coloneqq \left[ \vect{C}^{\intercal}, (\vect{CA})^{\intercal},
    \ldots,(\vect{CA}^{m-1})^{\intercal} \right]^{\intercal} \in
  \Real^{mN\times\rho}$. 
  
  With regards to a probability space, if
  \begin{enumerate*}[label = \textbf{(\roman*)}]

  \item $(\bm{\upsilon}_t)_t$ and
    $(\bm{\omega}_t)_t$ in \eqref{KARMA} are considered to be zero-mean,
    independent and identically distributed stochastic processes, as well as
    independent of each other,

  \item $(\bm{\omega}_t)_t$ is independent of $(\bm{\psi}_t)_t$, and

  \item $\bm{\omega}_t$ and $\bm{\psi}_{t'}$, $\forall (t, t')$ s.t.\ $t > t'$,
    are independent,
  \end{enumerate*}
  then
  \begin{align}
    \Expect\Set*{\tfrac{1}{\tau_{\text{f}}} \bm{\mathcal{F}}_{t+1} \Kprod
    \bm{\mathcal{B}}_t^{\intercal} \given \{ \bm{\psi}_{t'} \}_{t'=t-
    \tau_{\text{b}}+1}^{t+\tau_{\text{f}} + m -1} }  = \vect{O} \bm{\Pi}_{t+1}
    \,. \label{cond.expectation} 
  \end{align}

  If, in addition,
  \begin{enumerate*}[label = \textbf{(\roman*)}, resume*]
  \item $(\bm{\omega}_t)_t$, $(\bm{\upsilon}_t)_t$,
    $(\bm{\psi}_t)_t$, and
    $(\bm{\omega}_t \Kprod \bm{\psi}_{t-\tau}^{\intercal})_t$,
    $\forall \tau\in\IntegerPP$, are wide-sense stationary,
  \end{enumerate*}
  then
  $\lim_{\tau_{\text{f}} \to \infty} \bm{\mathcal{E}}_t^{\tau_{\text{f}}} =
  \vect{0}$, $\forall t$, in the mean-square ($\mathcal{L}_2$-) sense w.r.t.\
  the probability space.
  
\end{prop}


\begin{proofof}{Proposition \ref{Prop:O}}
  See \ref{App:prove.prop}.
\end{proofof}

Motivated by~\eqref{low.rank.formula} and \eqref{cond.expectation}, the result
($\lim_{\tau_{\text{f}} \to \infty} \bm{\mathcal{E}}_t^{\tau_{\text{f}}} =
\vect{0}$, $\forall t$), and the fact that the conditional expectation is the
least-squares-best estimator~\citep[\S9.4]{Williams.book}, the following task is
proposed to obtain an estimate of the observability matrix:
\begin{align}
  \left(\prescript{}{\mathcal{V}}{\hat{\vect{O}}}_t, \hat{\bm{\Pi}}_t \right) \in
  \Argmin_{\substack{\vect{O}\in\Real^{mN \times \rho}\\
  \bm{\Pi}\in\Real^{\rho\times \tau_{\text{b}} N}}} \norm*{\tfrac{1}{\tau_{\text{f}}}
  \bm{\mathcal{F}}_{t+1} \Kprod \bm{\mathcal{B}}_t^{\intercal} - \vect{O}
  \bm{\Pi}}_{\text{F}}^2 \,. \label{estimate.O}
\end{align}
To solve \eqref{estimate.O}, the singular value decomposition (SVD) is applied
to obtain
$(1/\tau_{\text{f}}) \bm{\mathcal{F}}_{t+1} \Kprod
\bm{\mathcal{B}}_t^{\intercal} = \vect{U} \bm{\Sigma}\vect{V}^{\intercal}$,
where $\vect{U}\in\mathbb{R}^{mN\times mN}$ is orthogonal. Assuming that
$\rho \leq \rank [(1/\tau_{\text{f}}) \bm{\mathcal{F}}_{t+1} \Kprod
\bm{\mathcal{B}}_t^{\intercal}]$, the Schmidt-Mirsky-Eckart-Young
theorem~\citep{ben2003generalized} provides the estimates
$\prescript{}{\mathcal{V}}{\hat{\vect{O}}}_t \coloneqq \vect{U}_{:,1:\rho}$ and
$\hat{\bm{\Pi}}_t \coloneqq \bm{\Sigma}_{1:\rho, 1:\rho}
\vect{V}^{\intercal}_{:,1:\rho}$, where $\vect{U}_{:,1:\rho}$ is the orthogonal
matrix that collects those columns of $\vect{U}$ that correspond to the top
(principal) $\rho$ singular values in $\bm{\Sigma}$.

Due to the factorization $\vect{O} \bm{\Pi}$, identifying the observability
matrix becomes ambiguous, since for any non-singular matrix
$\vect{P}\in \Real^{\rho\times \rho}$,
$\vect{O} \bm{\Pi} = \vect{O}\vect{P} \cdot \vect{P}^{-1} \bm{\Pi}$, and
$\prescript{}{\mathcal{V}}{\hat{\vect{O}}}_t \vect{P}$ can serve also as an
estimate. By virtue of the elementary observation that the column (range) spaces
of $\prescript{}{\mathcal{V}}{\hat{\vect{O}}}_t \vect{P}$ and
$\prescript{}{\mathcal{V}}{\hat{\vect{O}}}_t$ coincide, it becomes preferable to
identify the column space of $\prescript{}{\mathcal{V}}{\hat{\vect{O}}}_t$,
denoted hereafter by $[\prescript{}{\mathcal{V}}{\hat{\vect{O}}}_t]$, rather
than the matrix $\prescript{}{\mathcal{V}}{\hat{\vect{O}}}_t$ itself. If
$\rho = \rank [\prescript{}{\mathcal{V}}{\hat{\vect{O}}}_t]$, then
$[\prescript{}{\mathcal{V}}{\hat{\vect{O}}}_t]$ becomes a point in the Grassmann
manifold $\text{Gr}(\rho, mN)$, or Grassmannian, which is defined as the
collection of all linear subspaces of $\Real^{mN}$ with rank equal to
$\rho$~\citep[p.~73]{Loring.Tu.book}. The Grassmannian $\text{Gr}(\rho, mN)$ is a
Riemannian manifold with dimension equal to
$\rho(mN-\rho)$~\citep[p.~74]{Loring.Tu.book}. The algorithmic procedure of
extracting the feature $[\prescript{}{\mathcal{V}}{\hat{\vect{O}}}_t]$ from the
available data is summarized in Alg.~\ref{Algo:Map.O.to.Grassmannian}. To keep
notation as general as possible, instead of using all of the signal samples, a
subset $\mathfrak{T}\subset \Integer$ is considered and signal samples are
gathered in $(\prescript{}{\nu}{y}_{t})_{t\in \mathfrak{T}}$ per node $\nu$. 

\begin{algorithm}[!t]
  \DontPrintSemicolon
  \SetKwInOut{parameters}{Parameters}
  \SetKwInOut{input}{Input}
  \SetKwInOut{output}{Output}
  \SetKwBlock{initial}{Initialization}{}

  \parameters{Time indices $\mathfrak{T}$, and positive integers $N$, $m$,
    $\rho$, $\tau_{\text{f}}$, $\tau_{\text{b}}$.}

  \input{Time series $(\prescript{}{\mathcal{V}}{\vect{y}}_{t})_{t\in
        \mathfrak{T}}$.}
  
  \output{Grassmannian features $\Set{x_t}_{t\in\mathfrak{T}}$.}

  \BlankLine
  
    \For{all $t\in\mathfrak{T}$}{%

      Form
      $(1/\tau_{\text{f}}) \bm{\mathcal{F}}_{t+1} \Kprod
      \bm{\mathcal{B}}_t^{\intercal}$ via \eqref{FB}.

      Apply SVD:
      $(1/\tau_{\text{f}}) \bm{\mathcal{F}}_{t+1} \Kprod
      \bm{\mathcal{B}}_t^{\intercal} = \vect{U}
      \bm{\Sigma}\vect{V}^{\intercal}$.

      Feature
      $x_t \coloneqq [\prescript{}{\mathcal{V}}{\hat{\vect{O}}}_t] \in
      \text{Gr}(\rho, mN)$ is the linear subspace spanned by the $\rho$
      ``principal'' columns of $\vect{U}$.

    }
  
  \caption{Extracting Grassmannian features}\label{Algo:Map.O.to.Grassmannian}
\end{algorithm}

There can be many choices for the reproducing kernel function
$\kappa(\cdot, \cdot)$ (\cf Sec.~\ref{App:RKHS}). If the linear kernel
$\kappa_{\text{lin}}$ is chosen, then $\mathcal{H} = \Real^q$, $\varphi(\cdot)$
becomes the identity mapping,
$\bm{\varphi}_t = [\vect{y}_t^{\intercal}, \vect{y}_{t+1}^{\intercal}, \ldots,
\vect{y}_{t+N-1}^{\intercal}]^{\intercal} \in \Real^{qN}$, and $\Kprod$ boils
down to the usual matrix product. This case was introduced
in~\citep{slavakis2018clustering}. The most popular choice for $\kappa$ is the
Gaussian kernel $\kappa_{\text{G}; \sigma}$, where parameter $\sigma > 0$ stands
for standard deviation. However, pinpointing the appropriate $\sigma_*$ for a
specific dataset is a difficult task which may entail cumbersome
cross-validation procedures~\citep{Scholkopf.Smola.Book}. A popular approach to
circumvent the judicious selection of $\sigma_*$ is to use a dictionary of
parameters $\Set{\sigma_j}_{j=1}^J$, with $J\in\IntegerPP$, to cover an interval
where $\sigma_*$ is known to belong to. A reproducing kernel function
$\kappa(\cdot, \cdot)$ can be then defined as the convex combination
$\kappa(\cdot, \cdot) \coloneqq \sum_{j=1}^J \gamma_j \kappa_{\text{G};
  \sigma_j}(\cdot, \cdot)$, where $\Set{\gamma_j}_{j=1}^J$ are convex weights,
\ie, non-negative real numbers s.t.\
$\sum_{j=1}^J \gamma_j = 1$~\citep{Scholkopf.Smola.Book}. Such a strategy is
followed in Section~\ref{Sec:Tests}. Examples of non-Gaussian kernels
can be also found in Sec.~\ref{App:RKHS}.

Parameters in Alg.~\ref{Algo:Map.O.to.Grassmannian} need to be chosen properly
to guarantee that features $\Set{x_i}_{i\in\mathfrak{I}}$ capture the
statistical information of the time series. Parameters $N$, $m$ and $\rho$
control the dimension $\rho(mN-\rho)$ of the Grassmannian, which should be large
enough to capture the variability of the assumed low-dimensional feature
point-cloud. The sum $m + \tau_{\text{f}} + \tau_{\text{b}}$ should not be
greater than the length of the time series due to the size of "forward" and
"backward' matrices $\bm{\mathcal{F}}_{t}$ and $\bm{\mathcal{B}}_{t}$, while
large values of $\tau_{\text{f}}$ can help in reducing the estimation error of
${\prescript{}{\mathcal{V}}{\hat{\vect{O}}}}_t$.

\section{Network Clustering In The Grassmannian}\label{Sec:Clustering}

\subsection{Extended Geodesic Clustering by Tangent Spaces}

Having extracted and mapped features into the Grassmannian, the next task in the
pipeline of the framework is clustering. To keep this module as generic as
possible, the index set $\mathfrak{I}$ will be used henceforth to mark
features in $\Set{x_i}_{i\in\mathfrak{I}}$.

\begin{figure}[!t]
  \subfloat[Clusters on $\text{Gr}(\rho,
  mN)$]{\includegraphics[width=.4\linewidth]{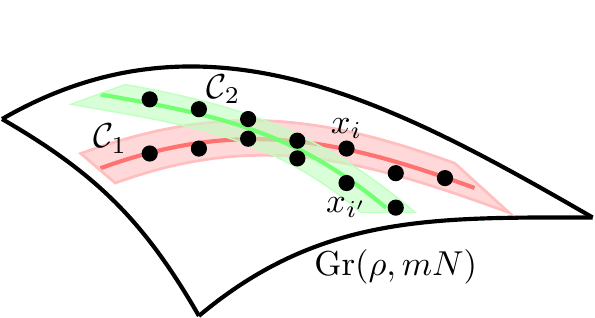}\label{Fig:RMMM}}
  \subfloat[Angular information]{\includegraphics[width =
    .6\linewidth]{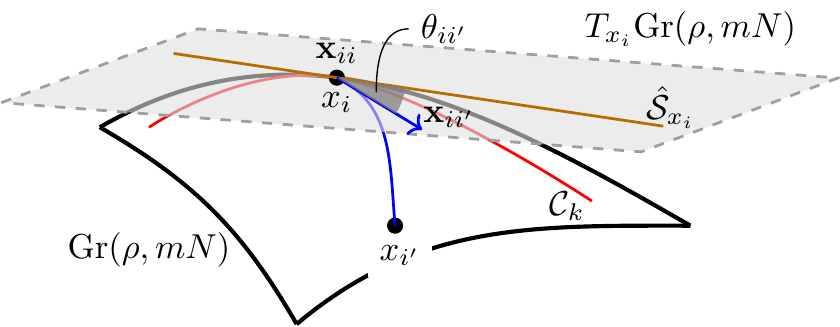}\label{Fig:theta}} 
  \caption{(a) The Riemannian multi-manifold modeling (RMMM) hypothesis. (b)
    Angular information computed in tangent spaces and used in
    Alg.~\ref{Algo:OnGrass}.}\label{Fig:RMMM.theta}
\end{figure}

This work follows the Riemannian multi-manifold modeling (RMMM)
hypothesis~\citep{RMMM.arxiv, RMMM.AISTATS.15, slavakis2018clustering}, where
clusters $\Set{\mathcal{C}_k}_{k=1}^K$ are considered to be submanifolds of the
Grassmannian, and data $\Set{x_i}_i$ are located close to or onto
$\Set{\mathcal{C}_k}_{k=1}^K$ (see Fig.~\ref{Fig:RMMM} for the case of $K=2$
clusters). RMMM allows for clusters to intersect; a case where the classical
K-means, for example, is known to face difficulties \citep{yang2011multi}.

Clustering is performed by Alg.~\ref{Algo:OnGrass}, coined geodesic clustering
by tangent spaces (GCT). The present GCT extends its initial form of
\citep{RMMM.arxiv, RMMM.AISTATS.15, slavakis2018clustering} to the case of
Alg.~\ref{Algo:OnGrass} where there is \textit{no}\/ need to know the number $K$
of clusters a-priori. This desirable feature of Alg.~\ref{Algo:OnGrass} is also
along the lines of usual practice, where it is unrealistic to know $K$ before
employing a clustering algorithm.

In a nutshell, Alg.~\ref{Algo:OnGrass} computes the affinity matrix $\vect{W}$
of features $\Set{x_i}_{i\in\mathfrak{I}}$ in
step~\ref{Step:OnGrass:adjacency.matrix}, comprising information about sparse
data approximations, via weights $\Set{\alpha_{ii'}}_{i, i'\in\mathfrak{I}}$, as
well as the angular information $\Set{\theta_{ii'}}_{i,
  i'\in\mathfrak{I}}$. Although the incorporation of sparse weights originates
from \citep{elhamifar2011sparse}, one of the novelties of GCT is the usage of the
angular information via $\Set{\theta_{ii'}}_{i, i'\in\mathfrak{I}}$. GCT's
version of~\citep{RMMM.arxiv, RMMM.AISTATS.15, slavakis2018clustering} applies
spectral clustering in step~\ref{Step:OnGrass:Louvain}, where knowledge of the
number of clusters $K$ is necessary. To surmount the obstacle of knowing $K$
beforehand, Louvain clustering method~\citep{aynaud2010static} is adopted in
step~\ref{Step:OnGrass:Louvain}. The Louvain method belongs to the family of
hierarchical-clustering algorithms that attempt to maximize a modularity
function, which monitors the intra- and inter-cluster density of
links/edges. Needless to say that any other hierarchical-clustering scheme can
be used at step~\ref{Step:OnGrass:Louvain} instead of Louvain method.

\begin{algorithm}[!t]
  \DontPrintSemicolon
  \SetKwInOut{input}{Input}
  \SetKwInOut{parameters}{Parameters}
  \SetKwInOut{output}{Output}
  \SetKwBlock{initial}{Initialization}{}

  \input{Grassmannian features $\Set{x_i}_{i\in \mathfrak{I}}$.}
  
  \parameters{$K_{\text{NN}}\in \IntegerPP$ and
    $\sigma_{\alpha}, \sigma_{\theta}\in \RealPP$.}
  
  \output{Clusters $\Set{\mathcal{C}_k}_{k=1}^K$.}

  \BlankLine

  \For{all $i\in\mathfrak{I}$\label{Step:OnGrass:every.datum}}
  {%

    Define the $K_{\text{NN}}$-nearest-neighbors
    $\mathscr{N}_{\text{NN}}(x_i)$. \label{Step:OnGrass:knn}
    
    Map $\mathscr{N}_{\text{NN}}(x_i)$ into the tangent space
    $T_{x_i}\text{Gr}(\rho, mN)$ of the Grassmannian at $x_i$ via the logarithm
    map: $\vect{x}_{ii'}\coloneqq \log_{x_i}(x_{i'})$,
    $\forall x_{i'}\in \mathscr{N}_{\text{NN}}(x_i)$. \label{Step:OnGrass:log}

    Identify $\Set{\alpha_{ii'}}_{x_{i'}\in \mathscr{N}_{\text{NN}}(x_i)}$ via
    \eqref{sparse.coding}. Set $\alpha_{ii'} \coloneqq 0$, for all $i'$ s.t.\
    $x_{i'}\notin \mathscr{N}_{\text{NN}}(x_i)$.\label{Step:OnGrass:sparse.coding}

    Compute the sample correlation matrix
    $\hat{\vect{C}}_{x_i}$ in
    \eqref{sample.covariance.matrix}. \label{Step:OnGrass:sample.covariance.matrix} 

    Perform principal component analysis (PCA) on $\hat{\vect{C}}_{x_i}$ to
    extract the eigenspace $\hat{\mathcal{S}}_{x_i}$. \label{Step:OnGrass:PCA}

    Compute angle $\theta_{ii'}$ between vector $\vect{x}_{ii'} - \vect{x}_{ii}$
    and $\hat{\mathcal{S}}_{x_i}$, $\forall x_{i'}\in \mathscr{N}_{\text{NN}}(x_i)$
    (${\theta_{ii}} \coloneqq 0$). Let also $\theta_{ii'} \coloneqq 0$ for
    $x_{i'}\notin \mathscr{N}_{\text{NN}}(x_i)$. \label{Step:OnGrass:theta}

  }

  Form the symmetric
  $\lvert \mathfrak{I}\rvert \times \lvert \mathfrak{I}\rvert$ affinity
  (adjacency) matrix $\vect{W} \coloneqq [w_{ii'}]_{(i,i')\in \mathfrak{I}^2}$,
  where entry $w_{ii'}$ is defined as \label{Step:OnGrass:adjacency.matrix}
  \[
    w_{ii'} \coloneqq \exp(\lvert \alpha_{ii'}\rvert + \lvert \alpha_{i'i}\rvert) \cdot
    \exp[-(\theta_{ii'} + \theta_{i'i})/\sigma_{\theta}] \,.
  \] 

  Apply Louvain method~\citep{aynaud2010static} to $\vect{W}$ to parcellate the
  data $(x_i)_{i\in \mathfrak{I}}$ into clusters
  $\Set{\mathcal{C}_k}_{k=1}^K$. \label{Step:OnGrass:Louvain}

  \caption{Extended geodesic clustering by tangent spaces (eGCT)}\label{Algo:OnGrass}
\end{algorithm}

A short description of the steps in Alg.~\ref{Algo:OnGrass} follows, with
Riemannian-geometry details deferred to \citep{RMMM.arxiv, RMMM.AISTATS.15,
  slavakis2018clustering}. Alg.~\ref{Algo:OnGrass} visits
$\Set{x_i}_{i\in\mathfrak{I}}$ sequentially
(step~\ref{Step:OnGrass:every.datum}). At step~\ref{Step:OnGrass:knn}, the
$K_{\text{NN}}$-nearest-neighbors $\mathscr{N}_{\text{NN}}(x_i)$ of $x_i$ are
identified, \ie, those $K_{\text{NN}}$ points, taken from $\Set{x_i}_i$, which
are placed the closest from $x_i$ with respect to the Grassmannian
distance~\citep{Absil.Gr.04}. The neighbors $\mathscr{N}_{\text{NN}}(x_i)$ are
then mapped at step~\ref{Step:OnGrass:log} to the Euclidean vectors
$\Set{\vect{x}_{ii'}}_{x_{i'}\in \mathscr{N}_{\text{NN}}(x_i)}$ in the tangent
space $T_{x_i}\text{Gr}(\rho, mN)$ of the Grassmannian at $x_i$ (the
gray-colored plane in Fig.~\ref{Fig:theta}) via the logarithm map
$\log_{x_i}(\cdot)$, whose computation (non-closed form via SVD) is provided in
\citep{RMMM.arxiv, slavakis2018clustering}. Step~\ref{Step:OnGrass:sparse.coding}
computes the weights
$\Set{\alpha_{ii'}}_{x_{i'} \in \mathscr{N}_{\text{NN}}(x_i)}$, with
$\alpha_{ii} \coloneqq 0$, via the following sparse-coding task:
\begin{align}
  \min_{\Set{\alpha_{ii'}}}
  {} & {} \norm*{\mathbf{x}_{ii}-\sum\nolimits_{x_{i'} \in \mathscr{N}_{\text{NN}}(x_i) \setminus \Set{x_i}}
       \alpha_{ii'}\mathbf{x}_{ii'}}^2 \notag\\  
  {} & {} + \sum\nolimits_{x_{i'} \in \mathscr{N}_{\text{NN}}(x_i) \setminus \Set{x_i}}
       \exp[{\norm{\mathbf{x}_{ii'}-\mathbf{x}_{ii}}/ \sigma_{\alpha}}]
       \cdot |\alpha_{ii'}| \notag\\ 
  \text{s.to}\ {}
     & {} \sum\nolimits_{x_{i'} \in \mathscr{N}_{\text{NN}}(x_i) \setminus \Set{x_i}}
       \alpha_{ii'} = 1 \,. \label{sparse.coding}
\end{align}

The affine constraint in \eqref{sparse.coding}, imposed on the
$\Set{\alpha_{ii'}}$ coefficients in representing $\mathbf{x}_{ii}$ via its
neighbors, is motivated by the affine nature of the tangent space
(Fig.~\ref{Fig:theta}). Moreover, the larger the distance of neighbor
$\mathbf{x}_{ii'}$ from $\mathbf{x}_{ii}$, the larger the weight
$\exp[{\norm{\mathbf{x}_{ii'} - \mathbf{x}_{ii}}/ \sigma_{\alpha}}]$, which in
turn penalizes severely the coefficient $\alpha_{ii'}$ by pushing it to values
close to zero. Step~\ref{Step:OnGrass:sample.covariance.matrix} computes the
sample covariance matrix
\begin{align}
  \hat{\vect{C}}_{x_i} \coloneqq \tfrac{1}{\lvert \mathscr{N}_{\text{NN}}(x_i)\rvert - 1}
  \sum\nolimits_{x_{i'}\in \mathscr{N}_{\text{NN}}(x_i)} (\vect{x}_{ii'} - \bar{\vect{x}}_i)
  (\vect{x}_{ii'} - \bar{\vect{x}}_i)^{\intercal}
  \,, \label{sample.covariance.matrix}
\end{align}
where
$\bar{\vect{x}}_i \coloneqq (1/\lvert \mathscr{N}_{\text{NN}}(x_i)\rvert)
\sum_{x_{i'}\in \mathscr{N}_{\text{NN}}(x_i)} \vect{x}_{ii'}$ denotes the sample
average of the neighbors of $\vect{x}_{ii}$. PCA is applied to
$\hat{\vect{C}}_{x_i}$ at step~\ref{Step:OnGrass:PCA} to compute the principal
eigenspace $\hat{S}_{x_i}$, which may be viewed as an approximation of the image
of the cluster (submanifold) $\mathcal{C}_k$, via the logarithm map, into the
tangent space $T_{x_i} \text{Gr}(\rho, mN)$ (see Fig.~\ref{Fig:theta}). Once
$\hat{S}_{x_i}$ is computed, the angle $\theta_{ii'}$ between vector
$\vect{x}_{ii'} - \vect{x}_{ii}$ and $\hat{\mathcal{S}}_{x_i}$ is also computed
at step~\ref{Step:OnGrass:theta} to extract angular information. The larger the
angle $\theta_{ii'}$ is, the less the likelihood for $x_{i'}$ to belong to
cluster $\mathcal{C}_k$. The additional use of angular information by GCT
advances the boundary of state-of-the-art clustering methods in the
Grassmannian, where, usually, the weights of the adjacency matrix are defined
via the Grassmannian (geodesic) distance or sparse-coding
schemes~\citep{elhamifar2011sparse}.

\subsection{Summarizing the Network-Clustering  Framework} \label{Sec:Summarize.framework}

To summarize, the flowchart of the network-clustering framework is presented in
Fig.~\ref{Fig:Flowchart}. The most straightforward path is the (blue-colored)
state-clustering one, where data are firstly formed (St1), then
Alg.~\ref{Algo:Map.O.to.Grassmannian} is applied to those data to collect
features (St2), and finally Alg.~\ref{Algo:OnGrass} is utilized to assign those
features into clusters $\Set{\mathcal{C}_k}_{k=1}^K$ (St3). In this context,
clustering is equivalent to parcellating the time horizon $\mathfrak{T}$ into a
partition $\Set{\mathfrak{T}_j}_{j=1}^J$ of time intervals s.t.\ data
$(\prescript{}{\lvert \mathcal{N} \rvert}{y}_{t})_{t\in\mathfrak{T}_j}$ are
mapped to the same state $j$.

The ``community-detection'' (red color) and
``subnetwork-state-sequence-clustering'' (green color) paths require state
clustering as a pre-processing part. This is necessary in order to achieve high
accuracy clustering results. Without knowing the starting and ending points of
different states, there will be time-series vectors
$\prescript{}{\nu}{\vect{y}}_{t}$ in Alg.~\ref{Algo:Map.O.to.Grassmannian} which
capture data from two consecutive states, since
$\prescript{}{\nu}{\vect{y}}_{t}$ takes the form of
$[\prescript{}{\nu}{y}_t, \ldots, \prescript{}{\nu}{y}_{t+ \texttt{Buff}_{\nu}
  -1}]^{\intercal}$. Features corresponding to those vectors will decrease the
clustering accuracy since the extracted features do not correspond to any actual
state or community. Once states are determined, the features that come from two consecutive states are ignored and the time horizon
$\mathfrak{T}$ is partitioned in $\Set{\mathfrak{T}_j}_{j=1}^J$, then
Algs.~\ref{Algo:Map.O.to.Grassmannian} and \ref{Algo:OnGrass} are applied per
state $j$ to detect communities (Comm4--Comm6). In ``subnetwork-state-sequence
clustering,'' states are again identified first. Per state, nodal time-series
data are formed according to Sec.~\ref{Sec:comm.task.clustering} (Sub4) and
nodal features are extracted by Alg.~\ref{Algo:Map.O.to.Grassmannian}
(Sub5). All those features from all states are collected and finally
Alg.~\ref{Algo:OnGrass} is applied to track/identify subnetwork state sequences
(Sub6).

\subsection{Computational Complexity}

The main computational burden comes from the feature extracting and clustering
steps in Alg.~\ref{Algo:Map.O.to.Grassmannian} and Alg.~\ref{Algo:OnGrass}. If
$\mathfrak{I}$ denotes the points in the Grassmannian, the computational
complexity for computing features $\Set{x_i}_{i\in\mathfrak{I}}$ in
Alg.~\ref{Algo:Map.O.to.Grassmannian} is
$\mathcal{O}(|\mathfrak{I} |\mathcal{C}_{\Kprod})$, where $\mathcal{C}_{\Kprod}$
denotes the cost of computing
$\bm{\mathcal{F}}_{t+1} \Kprod \bm{\mathcal{B}}_t^{\intercal}$, which includes
SVD computations. In Alg.~\ref{Algo:OnGrass}, the complexity for computing the
$\mathscr{N}_{\text{NN}}(x_i)$ nearest neighbors of $x_i$ is
$\mathcal{O}(|\mathfrak{I} |\mathcal{C}_{\text{dist}} +
\mathscr{N}_{\text{NN}}\log|\mathfrak{I}|)$, where $\mathcal{C}_{\text{dist}}$
denotes the cost of computing the Riemannian distance between any two points,
and $\mathscr{N}_{\text{NN}}\log|\mathfrak{I}|$ refers to the cost of finding
the $\mathscr{N}_{\text{NN}}$ nearest neighbors of $x_i$.
Step~\ref{Step:OnGrass:sparse.coding} of Alg.~\ref{Algo:OnGrass} is a
sparsity-promoting optimization task of (6) and let $\mathcal{C}_{\text{SC}}$
denotes the complexity to solve it. Under
$\mathcal{M} \coloneqq \text{Gr}(\rho, mN)$, step~\ref{Step:OnGrass:PCA} of
Alg.~\ref{Algo:OnGrass} involves the computation of the eigenvectors of the
sample covariance matrix $\hat{\vect{C}}_{x_i}$, with complexity of
$\mathcal{O}(\dim\mathcal{M} + K_{\text{NN}}^3)$. In
step~\ref{Step:OnGrass:theta}, the complexity for computing empirical geodesic
angles is $\mathcal{O}[|\mathfrak{I}| (\mathcal{C}_{\log} + \dim\mathcal{M})]$,
where $\mathcal{C}_{\log}$ is the complexity of computing the logarithm map
$\log_{x_i}(\cdot)$; for details, see~\citep{slavakis2018clustering}. For the
last step of Alg.~\ref{Algo:OnGrass}, the exact complexity of Louvain method is
not known but the method seems to run in time
$\mathcal{O}(|\mathfrak{I} | \log|\mathfrak{I} |)$ with most of the
computational effort spent on modularity optimization at first level, since
modularity optimization is known to be NP-hard \citep{de2011generalized}. To
summarize, the complexity of Alg.~\ref{Algo:OnGrass} is
$\mathcal{O} [|\mathfrak{I}|^2(\mathcal{C}_{\text{dist}} + \mathcal{C}_{\log} +
\dim\mathcal{M}) + (K_{\text{NN}}+1)|\mathfrak{I}| \log|\mathfrak{I}| +
|\mathfrak{I}| (\dim\mathcal{M} + K_{\text{NN}}^3)]$.

\section{Numerical Tests}\label{Sec:Tests}

This section validates the proposed framework on synthetic and real data. Tags
eGCT[Sker] and eGCT[Mker] denote the proposed framework whenever a single and
multiple kernel functions are employed, respectively. In the case where the
linear kernel is used, the K-ARMA method boils down to the eGCT method
of~\citep{slavakis2018clustering}. Apart from the classical K-means, other
competing algorithms are:
\begin{enumerate*}[label = \textbf{(\roman*)}]

\item The sparse manifold clustering and embedding
  (SMCE)~\citep{elhamifar2011sparse};

\item interaction K-means with PCA (IKM-PCA)~\citep{vijay2015brain};

\item graph-shift-operator estimation (GOE)~\citep{Segarra:17} from the popular
  graph-signal-processing framework;

\item independent component analysis (ICA)~\citep{allen2014tracking,
    sockeel2016large};

\item multivariate Granger causality (MVGC)~\citep{barnett2014mvgc,
    duggento2018multivariate};

\item 3D-windowed tensor approach (3D-WTA)~\citep{al2017tensor}.

\end{enumerate*}
More details are given in Sec.~\ref{Sec:Supp:Competing Algorithms} of the
supplementary file to abide by the thirty-pages limit for new paper submissions
imposed by this journal. SMCE, 3D-WTA, ICA and the classical K-means will be
compared against proposed framework on state clustering. SMCE, IKM-PCA, 3D-WTA,
GOA, ICA, MVGC and K-means will be used in community detection. Since none of
IKM-PCA, GOA, MVGC and 3D-WTA can perform subnetwork-state-sequence clustering
across multiple states, only the results of proposed framework and SMCE are
reported. To ensure fair comparisons, the parameters of all methods were tuned
to reach optimal performance for every scenario at hand.

The evaluation of all methods was based on the following two criteria:
\begin{enumerate*}[label = \textbf{(\roman*)}]

\item Clustering accuracy, defined as the number of correctly clustered data
  points (ground-truth labels are known) over the total number of points;

\item normalized mutual information (NMI)~\citep{schutze2008introduction}; and


\end{enumerate*}
In what follows, every numerical value of the previous criteria is the uniform
average of $20$ independently performed tests for the particular scenario at
hand.

\subsection{Synthetic Data}\label{Sec:synthetic.fMRI}

\begin{figure}[!t]
  \centering
  \subfloat[State 1]{\includegraphics[width =
    .15\linewidth]{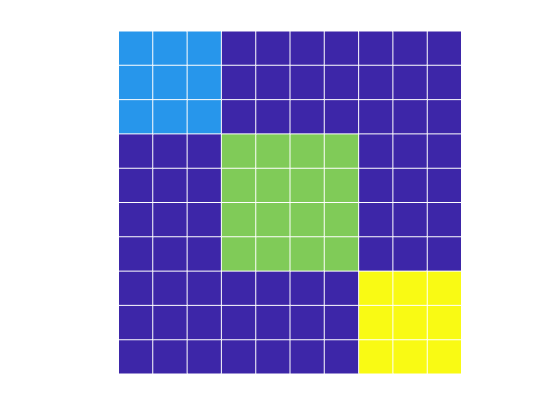}} 
  \subfloat[State 2]{\includegraphics[width =
    .15\linewidth]{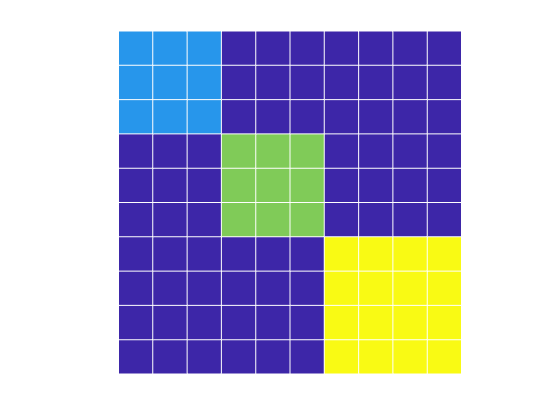}} 
  \subfloat[State 3]{\includegraphics[width =
    .15\linewidth]{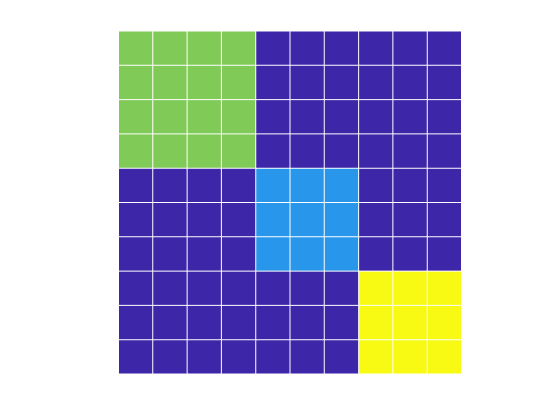}} 
  \subfloat[State 4]{\includegraphics[width =
    .15\linewidth]{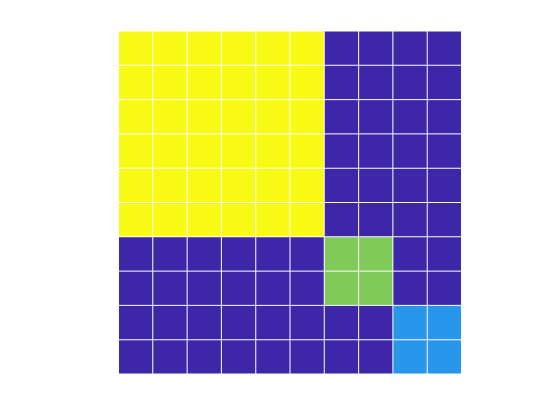}}\\ 
  \subfloat[BOLD time series of node \#2, dataset \#5]{\includegraphics[width =
    .6\linewidth]{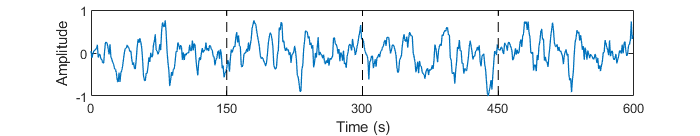}\label{Fig:fMRI.BOLD}}
  \caption{Synthetic data generated by the Matlab SimTB
    toolbox~\citep{allen2014tracking}. (a)-(d) Noiseless and outlier-free
    connectivity matrices corresponding to four network states. Nodes that share
    the same color cooperate to perform a common
    task.}\label{Fig:synthetic.fMRI}
\end{figure}

Data were generated by the open-source Matlab SimTB
toolbox~\citep{allen2014tracking}. A $10$-node network is considered that
transitions successively between $4$ distinct network states. Every state
corresponds to a certain connectivity matrix, generated via the following
path. Each connectivity matrix, fed to the SimTB toolbox, is modeled as the
superposition of three matrices: 1) The ground-truth (noiseless) connectivity
matrix (\cf Fig.~\ref{Fig:synthetic.fMRI}), where nodes sharing the same color
belong to the same cluster and collaborate to perform a common task; 2) a
symmetric matrix whose entries are drawn independently from a zero-mean Gaussian
distribution with standard deviation $\sigma$ to model noise; and 3) a symmetric
outlier matrix where $36$ entries are equal to $\mu$ to account for outlier
neural activity.

Different states may share different outlier matrices, controlled by
$\mu$. Aiming at extensive numerical tests, six datasets were generated
(corresponding to the columns of
Table~\ref{Table:synthetic.fMRI:state.clustering}) by choosing six pairs of
parameters $(\mu, \sigma)$ in the modeling of the connectivity matrices and the
SimTB toolbox. Datasets D1, D2 and D3 were created without outliers, while
datasets D4, D5 and D6 include outlier matrices with different $\mu$s in
different
states. Table~\ref{Table:parameters.of.synthetic.fMRI:state.clustering} details
the parameters of those six datasets. Driven by the previous connectivity
matrices, the SimTB toolbox generates BOLD time
series~\citep{ogawa1990brain}. Each state contributes $150$ signal samples, for a
total of $4\times 150 = 600$ samples, to every nodal time series, \eg
Fig.~\ref{Fig:fMRI.BOLD}.


\begin{table*}[tp]
  \centering
  \caption{Synthetic fMRI Data: State clustering}
  \label{Table:synthetic.fMRI:state.clustering}
  \resizebox{\linewidth}{!}{%
    \begin{threeparttable}
      \begin{tabular}{c|c|c|c||c|c|c||c|c|c||c|c|c}
        \toprule
        \multirow{3}{*}{Methods}
        & \multicolumn{6}{c}{Without Outliers}
        & \multicolumn{6}{c}{With Outliers}\cr
          \cmidrule(lr){2-7} \cmidrule(lr){8-13}
        & \multicolumn{3}{c}{Clustering Accuracy}
        & \multicolumn{3}{c}{NMI}
        & \multicolumn{3}{c}{Clustering Accuracy}
        & \multicolumn{3}{c}{NMI} \cr \cmidrule(lr){2-4} \cmidrule(lr){5-7}
          \cmidrule(lr){8-10}  
          \cmidrule(lr){11-13} %
              \cmidrule(lr){3-5} \cmidrule(lr){6-8} \cmidrule(lr){9-11} \cmidrule(lr){11-13}
        & D1 & D2 & D3 & D1 & D2 & D3 & D4 & D5 & D6 & D4 & D5 & D6 \cr\midrule
   eGCT & 0.969 & 0.805 & 0.640 & 0.948 & 0.766 & 0.596 & 0.944 & 0.743 & 0.589 & 0.860 & 0.627 & 0.340\cr
 eGCT[Sker] & \textbf{1} & 0.824 & 0.681 & \textbf{1} & 0.791 & 0.622 & 0.983 & 0.775 & 0.599 & 0.930 & 0.651 & 0.379\cr
 eGCT[Mker] & \textbf{1} & \textbf{0.839} & \textbf{0.708} & \textbf{1} & \textbf{0.808} & \textbf{0.641} & \textbf{0.992} & \textbf{0.800} & \textbf{0.626} & \textbf{0.967} & \textbf{0.689} &  \textbf{0.435}\cr 
 3DWTA~\citep{al2017tensor} & \textbf{1} & 0.792 & 0.603 & \textbf{1} & 0.735 & 0.556 & 0.943 & 0.731 & 0.517 & 0.872 & 0.562 & 0.281\cr              
 SMCE~\citep{elhamifar2011sparse} & 0.920 & 0.784 & 0.583 & 0.887 & 0.673 & 0.480 & 0.883 & 0.712 & 0.508 & 0.713 & 0.558 & 0.246\cr
 ICA~\citep{allen2014tracking,
    sockeel2016large} & 0.943 & 0.734 & 0.527 & 0.821 & 0.605 & 0.364 & 0.926 & 0.719 & 0.474 & 0.795 & 0.533 & 0.215 \cr
 Kmeans & 0.866 & 0.670 & 0.402 & 0.800 & 0.560 & 0.307 & 0.768 & 0.621 & 0.337 & 0.476 & 0.403 & 0.168\cr
 \bottomrule
      \end{tabular}
    \end{threeparttable}
  }
\end{table*}

Table~\ref{Table:synthetic.fMRI:state.clustering} demonstrates the results of
state clustering. The parameters used for eGCT, eGCT[Sker] and eGCT[Mker] are: $N \coloneqq 30$,
$m \coloneqq 2$, $\rho \coloneqq 2$, $\tau_{f} \coloneqq 60$,
$\tau_b \coloneqq 20$. The Gaussian kernel
$\kappa_{\text{G}; 0.8}(\cdot, \cdot)$ (\cf Sec.~\ref{App:RKHS}) is used in
the single-kernel method eGCT[Sker], while kernel
$\kappa(\cdot, \cdot) \coloneqq 0.6\,\kappa_{\text{G}; 0.8}(\cdot, \cdot) +
0.4\,\kappa_{\text{L}; 1}(\cdot, \cdot)$ is used in the eGCT[Mker] case since it
performed the best among other choices of kernel
functions. Fig.~\ref{Fig:synthetic.fMRI:state.clustering} depicts also the
standard deviations of the results of
Table~\ref{Table:synthetic.fMRI:state.clustering}, computed after performing
independent repetitions of the same test. To save space, the figures which
include the standard deviations of the subsequent tests will be omitted.

Among all methods, eGCT[Mker] scores the highest clustering accuracy and NMI over
all six datasets. It can be observed by
Table~\ref{Table:synthetic.fMRI:state.clustering} that the existence of outliers
affects negatively the ability of all methods to cluster data. The main reason
is that the algorithms tend to detect outliers and gather those in clusters
different from the nominal ones. Ways to reject those outliers are outside of
the scope of this study and will be provided in a future publication.

\begin{table*}[tp]
  \centering
  \caption{Synthetic fMRI data: Community detection}
  \label{Table:synthetic.fMRI:community.detection}
  \resizebox{\linewidth}{!}{%
    \begin{threeparttable}
      \begin{tabular}{c|c|c|c||c|c|c||c|c|c||c|c|c}
        \toprule
        \multirow{3}{*}{Methods}
        & \multicolumn{6}{c}{Without Outliers}
        & \multicolumn{6}{c}{With Outliers}\cr
          \cmidrule(lr){2-7} \cmidrule(lr){8-13}
        & \multicolumn{3}{c}{\begin{tabular}{@{}c@{}}Clustering\\ Accuracy\end{tabular}}
        & \multicolumn{3}{c}{NMI}
        & \multicolumn{3}{c}{\begin{tabular}{@{}c@{}}Clustering\\ Accuracy\end{tabular}}
        & \multicolumn{3}{c}{NMI}\cr
         \cmidrule(lr){2-4} \cmidrule(lr){5-7} \cmidrule(lr){8-10} \cmidrule(lr){11-13}
        & D1    & D2    &D3   & D1    & D2    &D3   & D4    & D5   &D6    & D4    & D5    &D6\cr\midrule
        eGCT & \textbf{1}  & 0.960 & 0.842 & \textbf{1}  & 0.876 & 0.775 & 0.973 & 0.910 & 0.817 & 0.940 & 0.793 & 0.664\cr
        eGCT[Sker]  & \textbf{1}  & \textbf{1}  & 0.915 & \textbf{1}  & \textbf{1}  & 0.838 & \textbf{1}  & 0.942 & 0.852 & \textbf{1}  & 0.864 & 0.710\cr
        eGCT[Mker]  & \textbf{1}  & \textbf{1}  & \textbf{0.945}  & \textbf{1} & \textbf{1}  & \textbf{0.907}  & \textbf{1} & \textbf{0.958} & \textbf{0.879}
        &\textbf{1} & \textbf{0.892} &\textbf{0.803}\cr
        3DWTA~\citep{al2017tensor} & \textbf{1}  & 0.951 & 0.839 & \textbf{1} & 0.927 & 0.754 & 0.925 & 0.863 & 0.799 & 0.842 & 0.780 & 0.638 \cr  
        SMCE~\citep{elhamifar2011sparse} & 0.965 & 0.929 & 0.827 & 0.902 & 0.865 & 0.691 & 0.909 & 0.773 & 0.745 & 0.769 & 0.647 & 0.563 \cr
        GOE~\citep{Segarra:17} & \textbf{1}  & 0.933 & 0.809 & \textbf{1} & 0.915 & 0.655 & 0.918 & 0.740 &0.684 &0.833 &0.652 &0.409 \cr
 ICA~\citep{allen2014tracking,
    sockeel2016large} & 0.974 & 0.936 & 0.830 & 0.917 & 0.883 & 0.702 & 0.910 & 0.826 & 0.761 & 0.828 & 0.715 & 0.592 \cr
 MVGC~\citep{barnett2014mvgc,
    duggento2018multivariate} & \textbf{1} & 0.948 & 0.834 & \textbf{1} & 0.920 & 0.722 & 0.914 & 0.845 & 0.759 & 0.826 & 0.742 & 0.611 \cr
 IKM-PCA~\citep{vijay2015brain} & 0.948 & 0.907 & 0.791 & 0.890 & 0.814 & 0.629 & 0.892 & 0.756 & 0.712 & 0.738 & 0.551 & 0.486\cr
 Kmeans & 0.908 & 0.876 & 0.725 & 0.810 & 0.729 & 0.547 & 0.843 & 0.672 & 0.605 & 0.620 & 0.391 & 0.314\cr
 \bottomrule
      \end{tabular}
      \end{threeparttable}
  }  
\end{table*}

Table~\ref{Table:synthetic.fMRI:community.detection} presents the results of
community detection. The numerical values in
Table~\ref{Table:synthetic.fMRI:community.detection} stand for the average
values over the $4$ states for each one of the datasets. Parameters of eGCT, eGCT[Sker] and eGCT[Mker] were set as follows: $N \coloneqq 30$,
$\texttt{Buff}_{\nu} \coloneqq 20$, $m \coloneqq 3$, $\rho \coloneqq 2$,
$\tau_f \coloneqq 50$, $\tau_b \coloneqq 10$. In eGCT[Sker], the utilized kernel
function is $\kappa_{\text{G}; 0.5}(\cdot, \cdot)$, while in eGCT[Mker] the
kernel is defined as
$\kappa(\cdot, \cdot) \coloneqq 0.5\, \kappa_{\text{G}; 0.5}(\cdot, \cdot) +
0.5\, \kappa_{\text{L}0; 1}(\cdot, \cdot)$ (\cf
Sec.~\ref{App:RKHS}). Table~\ref{Table:synthetic.fMRI:community.detection}
demonstrates that eGCT[Mker] consistently outperforms all other methods across
all datasets and even for the case where outliers contaminate the
data. Fig.~\ref{Fig:synthetic.fMRI:community detection} depicts also the
standard deviations of the results of
Table~\ref{Table:synthetic.fMRI:community.detection}.

\begin{table*}[h]
  \centering
  \caption{Synthetic fMRI data: Subnetwork state sequences}
  \label{Table:synthetic.fMRI:task.clustering}
  \resizebox{\linewidth}{!}{%
    \begin{threeparttable}
      \begin{tabular}{c|c|c|c||c|c|c||c|c|c||c|c|c}
        \toprule
        \multirow{3}{*}{Methods}
        & \multicolumn{6}{c}{Without Outliers}
        & \multicolumn{6}{c}{With Outliers}\cr
          \cmidrule(lr){2-7} \cmidrule(lr){8-13}
        & \multicolumn{3}{c}{\begin{tabular}{@{}c@{}}Clustering\\ Accuracy\end{tabular}}
        & \multicolumn{3}{c}{NMI}
        & \multicolumn{3}{c}{\begin{tabular}{@{}c@{}}Clustering\\ Accuracy\end{tabular}}
         & \multicolumn{3}{c}{NMI}\cr
         \cmidrule(lr){2-4} \cmidrule(lr){5-7} \cmidrule(lr){8-10} \cmidrule(lr){11-13}
         & D1    & D2    &D3   & D1    & D2    &D3   & D4    & D5   &D6    & D4    & D5    &D6\cr \midrule
   eGCT & \textbf{1}  & 0.816 & 0.749 & \textbf{1} & 0.767 & 0.684 & 0.928 & 0.701 & 0.633 & 0.874 & 0.484 & 0.355\cr
    eGCT[Sker] & \textbf{1} & 0.856 & 0.781 & \textbf{1} & 0.791 & 0.702 & 0.956 & 0.728 & 0.664 & 0.913 & 0.534 & 0.410\cr
    eGCT[Mker] & \textbf{1} & \textbf{0.884} & \textbf{0.817} & \textbf{1} & \textbf{0.821} & \textbf{0.739} & \textbf{1} & \textbf{0.757} & \textbf{0.721} & \textbf{1}  & \textbf{0.602} & \textbf{0.485}\cr
 SMCE~\citep{elhamifar2011sparse} & 0.936 & 0.792 & 0.691 & 0.804 & 0.617 & 0.495 & 0.851 & 0.665 & 0.580 & 0.785 & 0.416 & 0.318 \cr                                                                               \bottomrule
      \end{tabular}
       \end{threeparttable}
  }
\end{table*}

\begin{figure}[!t]
  \centering
  \subfloat[Datasets 1, 2, 3]{
  \includegraphics[width = .5\linewidth]{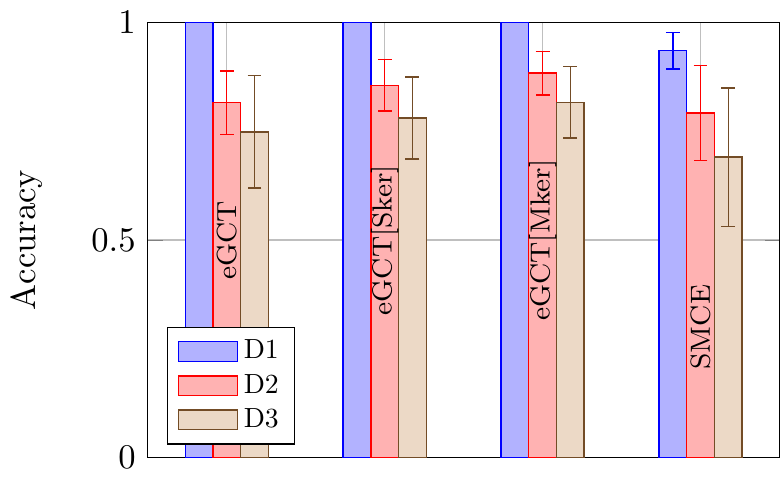}
  }
  
\subfloat[Datasets 4, 5, 6]{
\includegraphics[width = .5\columnwidth]{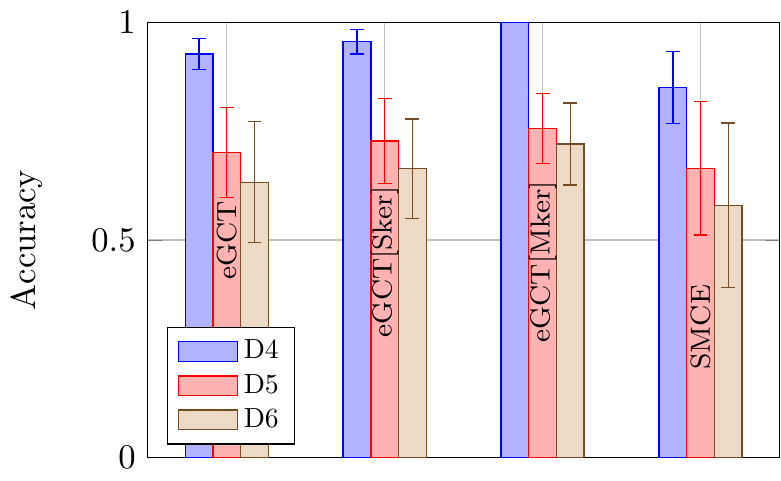}
    }
    \caption{Subnetwork-state-sequence clustering results of synthetic fMRI
      datasets. (a) Data without an independent event; (b) Data with an
      independent event. These data sets are the same as D1, D2, etc, in Table~\ref{Table:synthetic.fMRI:task.clustering}.} \label{Fig:synthetic.fMRI:task tracking}
\end{figure}

Table~\ref{Table:synthetic.fMRI:task.clustering} illustrates the results of
subnetwork-state-sequence clustering. The parameters of eGCT, eGCT[Sker] and eGCT[Mker]
were set as follows: $N \coloneqq 20$, $\texttt{Buff}_{\nu} \coloneqq 50$,
$m \coloneqq 3$, $\rho \coloneqq 3$, $\tau_f \coloneqq 45$,
$\tau_b \coloneqq 5$. The kernel functions used in eGCT[Sker] and eGCT[Mker] are
identical to those employed in
Table~\ref{Table:synthetic.fMRI:community.detection}. Similarly to the previous
cases, eGCT[Mker] outperforms all other methods across all datasets and scenarios
on both clustering accuracy and NMI. Fig.~\ref{Fig:synthetic.fMRI:task tracking}
depicts also the standard deviations of the results of
Table~\ref{Table:synthetic.fMRI:task.clustering}.

\subsection{Real Data}\label{Sec:real.data}

To validate the community detection framework,we tested our algorithm on functional networks derived for two subjects taken from the S1200 dataset of the Human
Connectome Project (HCP)~\citep{glasser2013minimal} were considered.

To avoid irrelevant influence, only the part of cleaned volume data in single
run with left-to-right phase encoding direction was employed. In addition to the
HCP preprocessing, each voxel was standardized by first subtracting the temporal
mean and then applying global signal regression. Specifically, motion outliers was used to estimate framewise displacement (FD) \citep{jenkinson2002improved} and volumes with FD>0.2 mm were censored and removed from further analysis.
In addition, we standardized each voxel by first subtracting the temporal mean and then applying global signal regression. Brain regions were defined using either the standard 116 region AAL-atlas \citep{tzourio2002automated}. The temporal activity for a given  brain region was computed by averaging the signal over all voxels within the region.

\begin{table}[t!]
  \centering
  \caption{Real fMRI data: community detection results}
  \label{Table:Real.fMRI:community detection}
  \begin{tabular}{c|c}
    \toprule
    Community & Fitting rate\cr\midrule
    Cerebellum & 0.381  \cr
    Control  & 0.440  \cr
    Default Mode & 0.642  \cr
    Dorsal Attention & 0.422 \cr
    Limbic & 0.386  \cr
    Salience/Ventral Attention & 0.700 \cr      
    Somatomotor & 0.554 \cr 
    Subcortical & 0.357 \cr 
    Visual & 0.633 \cr 
      \bottomrule
  \end{tabular}
\end{table}

\begin{figure}[t!]
  \centering

    \includegraphics[width = .5\columnwidth]{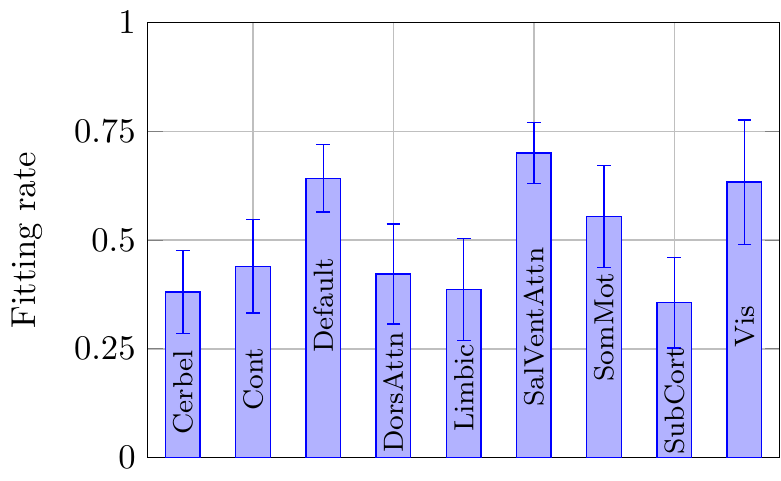}

  \caption{Real fMRI data : Community detection}
  \label{Fig:real.fMRI:community.detection}
\end{figure}

Table~\ref{Table:Real.fMRI:community detection} and Fig.~\ref{Fig:real.fMRI:community.detection} shows the community-detection results with 116 brain ROIs. Ten subjects are randomly selected from the HCP resting state fMRI data set. Nine cortical regions are considered as nine communities as labels of cognitive system. Each cortical region from the AAL atlas was mapped onto a cognitive system from the 7-Network parcellation scheme from the Schaefer-100 atlas, respectively \citep{schaefer2017local}. Community label assignment was based on minimizing the Euclidian distance from the centroid of a region in the AAL to the corresponding Schaefer-100atlas over more than 1000 samples. Subcortical and Cerebellar regions were combined into their respective systems. 

Table~\ref{Table:Real.fMRI:community detection} and Fig.~\ref{Fig:real.fMRI:community.detection} shows the community-detection results with 116 brain ROIs. Nodes/ROIs with the same color
are in the same cluster. Ten samples are randomly selected from the data set. .

The state clustering results of real fMRI are briefly described in
Sec.~\ref{Sec:Supp:Real.data} of the supplementary file.

\section{Supplementary: Competing Algorithms}\label{Sec:Supp:Competing Algorithms}

\subsection{Sparse Manifold Clustering and Embedding (SMCE)~\citep{elhamifar2011sparse}}

Each point on the Grassmannian is described by a sparse affine combination of
its neighbors. The computed sparse weights define the entries of a similarity
matrix, which is subsequently used to identify data-cluster associations. SMCE
does not utilize any angular information, as step~\ref{Step:OnGrass:theta} of
Alg.~\ref{Algo:OnGrass} does.

\subsection{Interaction K-means with PCA (IKM-PCA)~\citep{vijay2015brain}}

IKM is a clustering algorithm based on the classical K-means and Euclidean
distances within a properly chosen feature space. To promote time-efficient
solutions, the classical PCA is employed as a dimensionality-reduction tool for
feature-subset selection. In this algorithm, the dimension of fMRI data is reduced 
by classical PCA first, then the PCA-processed data are clustered using IKM. 

\subsection{Graph-shift-operator estimation (GOE)~\citep{Segarra:17}}

The graph shift operator is a symmetric matrix capturing the network's
structure, \ie topology. There are widely adopted choices of graph shift
operators, including the adjacency and Laplacian matrices, or their various
degree-normalized counterparts. An estimation algorithm in~\citep{Segarra:17}
computes the optimal graph shift operator via convex optimization. The computed
graph shift operator is fed to a spectral-clustering module to identify
communities within a single brain state, since \citep{Segarra:17} assumes
stationary time-series data.

\subsection{Independent Component Analysis based algorithms
  (ICA)~\citep{allen2014tracking, sockeel2016large}}

Independent component analysis discovers hidden features or factors from a set
of observed data such that the discovered features are maximally
independent. For state clustering, group ICA~\citep{allen2014tracking} is
introduced. In this algorithm, features are extracted and examined for
relationships among the data types at the group level (\ie, variations among
time sliding windows, patients or controls). Then, functional
connectivity matrices are estimated as covariance matrices and clustered by
K-means. For community detection, \citep{sockeel2016large} proposed a framework
with ICA and hierarchical clustering to identify functional brain connectivity
patterns of EEG and fMRI datasets.

\subsection{Multivariate Granger causality (MVGC)~\citep{barnett2014mvgc, duggento2018multivariate}}

To explore the knowledge of functional brain network as well as connectivity
patterns and community structures, multivariate Granger causality (MVGC) has
recently been applied to incorporate information about the influence exerted by
a brain region onto another. A MVGC toolbox is provided
by~\citep{barnett2014mvgc} that estimates ``Granger causality'' and vector
autoregressive coefficients on time or frequency domain of time series. A
community detection framework based on MVGC toolbox is proposed
in~\citep{duggento2018multivariate}. ``Granger causality'' strength between each
pair of nodes/ROIs become the entries of an adjacency matrix, which is fed into
spectral clustering for community detection.

\subsection{3D-Windowed Tensor Approach (3D-WTA)~\citep{al2017tensor}}

3D-WTA was originally introduced for community detection in dynamic networks by
applying tensor decompositions onto a sequence of adjacency matrices indexed
over the time axis. 3D-WTA was modified in~\citep{al2018tensor} to accommodate
multi-layer network structures. High-order SVD (HOSVD) and high-order orthogonal
iteration (HOOI) are used within a pre-defined sliding window to extract
subspace information from the adjacency matrices. The ``asymptotic-surprise''
metric is used as the criterion to determine the number of clusters. 3D-WTA is
capable of performing both state clustering and community detection.

\section{Supplementary: Synthetic fMRI data}\label{Sec:Supp:Synthetic.fMRI}

Table~\ref{Table:parameters.of.synthetic.fMRI:state.clustering} provides the
parameters $\mu$ and $\sigma$ used to generate noise matrices and symmetric
matrices to simulate outlier neural activities. By choosing different
combinations of $(\mu, \sigma)$, $6$ different synthetic fMRI datasets were
created.

\begin{table*}
\caption{Parameters $(\mu,\sigma)$ used to generate synthetic BOLD time series}
\label{Table:parameters.of.synthetic.fMRI:state.clustering}
\centering
\begin{tabu} to \hsize {|X|X|X|X|X|}
\hline
$Dataset$  &$State\ 1$  &$State\ 2$   &$State\ 3$  &$State\ 4$\\
\hline
1    &$(0,-10\text{dB})$ &$(0,-10\text{dB})$ &$(0,-10\text{dB})$ &$(0,-10\text{dB})$   \\
\hline
2    &$(0,-8\text{dB})$ &$(0,-8\text{dB})$ &$(0,-8\text{dB})$ &$(0,-8\text{dB})$   \\
\hline
3    &$(0,-6\text{dB})$ &$(0,-6\text{dB})$ &$(0,-6\text{dB})$ &$(0,-6\text{dB})$   \\
\hline
4    &$(0.2,-10\text{dB})$ &$(0.3,-10\text{dB})$ &$(0.4,-10\text{dB})$  &$(0.5,-10\text{dB})$   \\
\hline
5    &$(0.2,-8\text{dB})$ &$(0.3,-8\text{dB})$ &$(0.4,-8\text{dB})$ &$(0.5,-8\text{dB})$   \\
\hline
6    &$(0.2,-6\text{dB})$ &$(0.3,-6\text{dB})$ &$(0.4,-6\text{dB})$ &$(0.5,-6\text{dB})$   \\
\hline
\end{tabu}
\end{table*}

\begin{figure}[t!]
\centering
  \subfloat[Datasets 1, 2, 3]{
  \includegraphics[width = .5\linewidth]{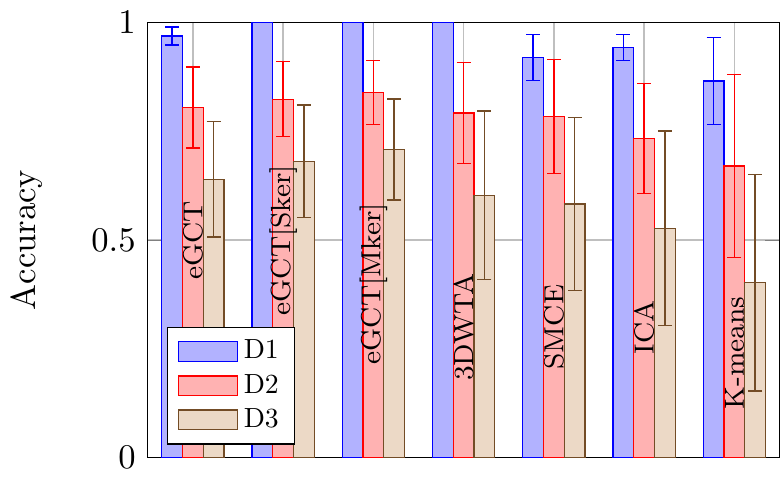}
     
  }

  \subfloat[Datasets 4, 5, 6]{
  \includegraphics[width = .5\linewidth]{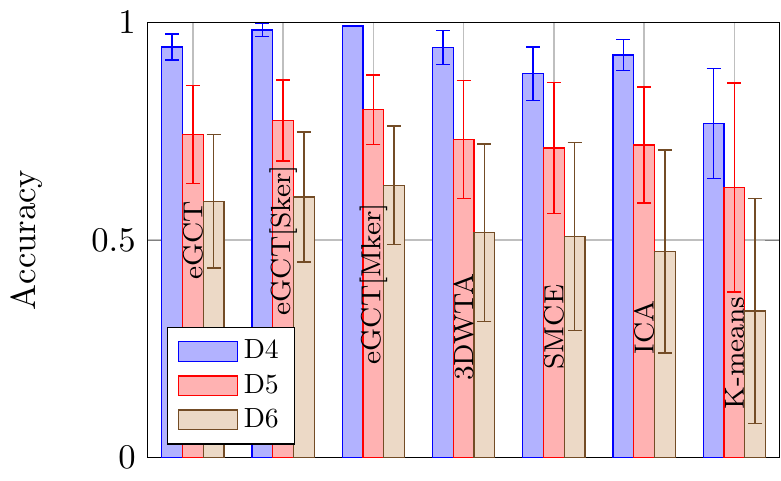}
  }
  \caption{State-clustering results of synthetic fMRI datasets. (a) Data without
    an independent event; (b) Data with an independent
    event. These data sets are the same as D1, D2, etc, in Table~\ref{Table:synthetic.fMRI:state.clustering}.} \label{Fig:synthetic.fMRI:state.clustering} 
\end{figure}

Standard-deviation results of state clustering on synthetic fMRI datasets are
demonstrated in Fig.~\ref{Fig:synthetic.fMRI:state.clustering}. Standard
deviation of all algorithms increase when the strength of the noisy matrix
increases. For dataset D1, eGCT[Sker] , eGCT[Mker] and 3DWTA reach $100\%$
accuracy; for other datasets, eGCT[Mker] exhibits the highest accuracy and the
smallest standard deviation.

\begin{figure}[t!]
  \centering
  \subfloat[Datasets 1, 2, 3]{
    \includegraphics[width = .5\linewidth]{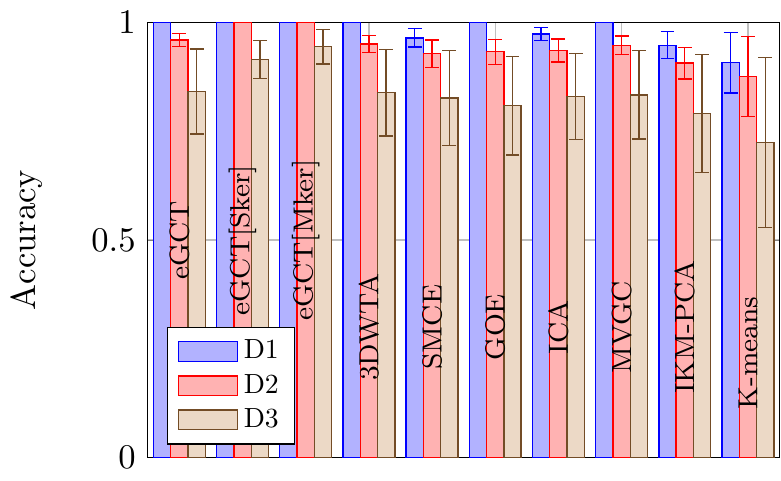}
  }
  
\subfloat[Datasets 4, 5, 6]{
 \includegraphics[width = .5\columnwidth]{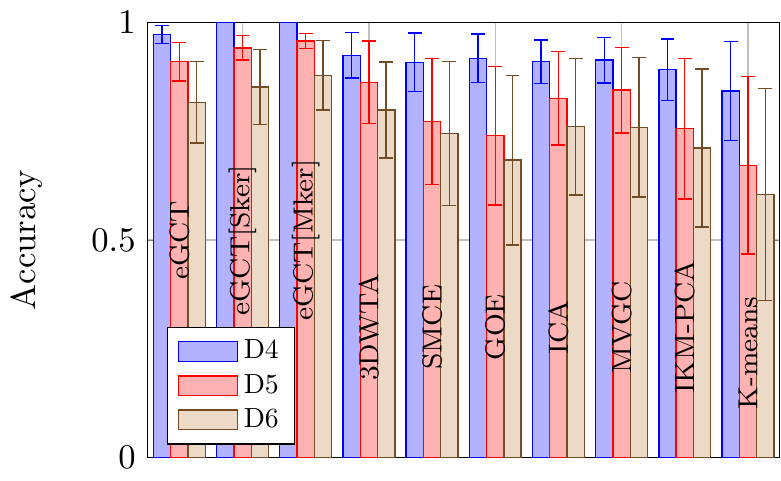}
     }
     \caption{Community detection results of synthetic fMRI datasets. (a) Data
      without an independent event; (b) Data with an independent
      event. These data sets are the same as D1, D2, etc, in Table~\ref{Table:synthetic.fMRI:community.detection}.} \label{Fig:synthetic.fMRI:community detection}
\end{figure}

Fig.~\ref{Fig:synthetic.fMRI:community detection} illustrates the results of
community detection for the synthetic fMRI datasets. eGCT, eGCT[Sker] , eGCT[Mker]
and 3DWTA score $100\%$ accuracy for dataset D1, while eGCT[Sker]  and eGCT[Mker]
show $100\%$ accuracy for dataset D4. eGCT[Mker] shows the highest accuracy on
all other datasets.

Standard-deviation results for subnetwork-state-sequence clustering on synthetic
fMRI datasets are demonstrated in Fig.~\ref{Fig:synthetic.fMRI:task
  tracking}. eGCT, eGCT[Sker]  and eGCT[Mker] score $100\%$ accuracy on dataset
D1. eGCT[Mker] shows the highest accuracy with the smallest standard deviation on
all other datasets.

\section{Supplementary: Real data}\label{Sec:Supp:Real.data}

Real fMRI behavioral data, acquired from the Stellar Chance 3T scanner (SC3T) at
the University of Pennsylvania, were used to cluster different states. The time
series in data are collected in two arms before and after an inhibitory sequence
of transcranial magnetic stimulation (TMS) known as continuous theta burst
stimulation~\citep{huang2005theta}. Real and Sham stimulation of two different
tasks were applied for TMS. The two behavioral tasks are: 1) Navon task: A big
shape made up of little shapes is shown on the screen. The big shape can either
be green or white in color. If green, participant identifies the big shape,
while if white, the participant identifies the little shape. The task was
presented in three blocks: All white stimuli, all green stimuli, and switching
between colors on 70\% of trials to introduce switching demands. Responses given
via button box are in the order of circle, x, triangle, square; 2) Stroop task:
Words are displayed in different color inks. There are two difficulty
conditions; one where subjects respond to words that introduced low color-word
conflict (far, deal, horse, plenty) or high conflict with color words differing
from the color the word is printed in (\eg red printed in blue, green printed
in yellow, etc.)~\citep{medaglia2018functional}. The participant has to tell the
color of the ink the word is printed in using a button box in the order of red,
green, yellow, blue.

Each BOLD time series was collected during an $8$min scan with
$\text{TR}=500$ms, which means that the length of time series is $956$. The time
series has $83$ cortical and subcortical regions so
$|\mathcal{N}| \coloneqq 83$. To test the state clustering results of fMRI time
series, 3 states are concatenated to create a single time series with length
$3\times 956 = 2,868$. The 3 states are: 1) Before real stimulation of the Navon
task; 2) after real stimulation of the Navon task; and 3) after real stimulation
of the Stroop task.

Parameters of eGCT, eGCT[Sker]  and eGCT[Mker] are defined as: $N \coloneqq 180$,
$m \coloneqq 4$, $\rho \coloneqq 2$, $\tau_f \coloneqq 350$,
$\tau_b \coloneqq 20$. In eGCT[Sker] , the kernel function is set equal to
$\kappa_{\text{G}; 0.45}(\cdot, \cdot)$, while in eGCT[Mker]
$\kappa(\cdot, \cdot) \coloneqq 0.3\, \kappa_{\text{G}; 0.25}(\cdot, \cdot) +
0.3\, \kappa_{\text{G}; 0.9}(\cdot, \cdot) + 0.4\, \kappa_{\text{L};0.75}(\cdot,
\cdot)$. Notice here that due to the sliding-window implementation in the
proposed framework, there are cases where the sliding window captures samples
from two consecutive states. 

\begin{table}[tp!]
  \centering
  \caption{Real fMRI data: State clustering results}
  \label{Table:Real.fMRI:state.clustering}
  \begin{tabular}{c|c|c}
    \toprule
    Methods & Clustering accuracy & NMI\cr\midrule
    eGCT & 0.885 & 0.809 \cr
    eGCT[Sker]  & 0.904 & 0.843 \cr
    eGCT[Mker] & \textbf{0.919} & \textbf{0.875} \cr
    SMCE & 0.893 & 0.816 \cr
    ICA &0.873 &0.776 \cr
    Kmeans & 0.801 & 0.720 \cr                           
      \bottomrule
  \end{tabular}
\end{table}

\begin{figure}[htpb!]
  \centering
  \includegraphics[width = .5\linewidth]{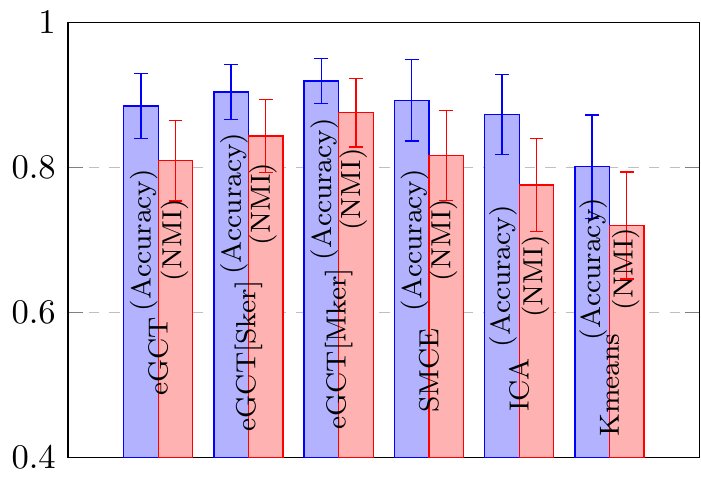}
  \caption{Real fMRI data: State clustering.}
  \label{Fig:real.fMRI:state.clustering}
\end{figure}

Results of state clustering on real fMRI data are revealed in
Table~\ref{Table:Real.fMRI:state.clustering}. Fig.~\ref{Fig:real.fMRI:state.clustering}
depicts also the standard deviations of the results of
Table~\ref{Table:Real.fMRI:state.clustering}. eGCT[Mker] scores the best
performance among all methods. 

\section{Conclusions}\label{Sec:conclusions}

This paper introduced a novel clustering framework to address all possible
clustering tasks in dynamic (brain) networks: state clustering, community
detection and subnetwork-state-sequence tracking/identification. Features were
extracted by a kernel-based ARMA model, with column spaces of observability
matrices mapped to the Grassmann manifold (Grassmannian). A clustering
algorithm, the geodesic clustering with tangent spaces, was also provided to
exploit the rich underlying Riemannian geometry of the Grassmannian, without the
need to know the number of clusters a-priori. The framework was validated on
multiple simulated and real datasets and compared against state-of-the-art
clustering algorithms. Test results demonstrate that the proposed framework
outperforms the competing methods in all clustering tasks. Current research
effort includes finding ways to reduce the size of the computational footprint
of the framework, and techniques to reject network-wide outlier data.

\section{Acknowledgements}

The brain network data \citep{glasser2013minimal} were provided by the Human
Connectome Project, MGH-USC Consortium (Principal Investigators: Bruce R.\ Rosen,
Arthur W.\ Toga and Van Wedeen; U01MH093765) funded by the NIH Blueprint
Initiative for Neuroscience Research grant; the National Institutes of Health
grant P41EB015896; and the Instrumentation Grants S10RR023043, 1S10RR023401,
1S10RR019307.

\appendix

\section{Reproducing Kernel Hilbert Spaces}\label{App:RKHS}

A reproducing kernel Hilbert space $\mathcal{H}$, equipped with inner product
$\innerp{\cdot}{\cdot}_{\mathcal{H}}$, is a functional space where each point
$g\in \mathcal{H}$ is a function
$g: \Real^q\to \Real: \vect{y} \mapsto g(\vect{y})$, for some $q\in\IntegerPP$,
s.t.\ the mapping $g\mapsto g(\vect{y})$ is continuous, for any choice of
$\vect{y}$~\citep{Aronszajn.50, Scholkopf.Smola.Book,
  Slavakis.OL.BookChapter.14}. There exists a kernel function
$\kappa(\cdot, \cdot): \Real^q\times \Real^q \to \Real$, unique to
$\mathcal{H}$, such that (s.t.)
$\varphi(\vect{y}) \coloneqq \kappa(\vect{y}, \cdot)\in \mathcal{H}$ and
$g(\vect{y}) = \innerp{g}{\varphi(\vect{y})}_{\mathcal{H}}$, for any
$g\in \mathcal{H}$ and any $\vect{y}\in \Real^q$~\citep{Aronszajn.50,
  Slavakis.OL.BookChapter.14}. The latter property is the reason for calling
kernel $\kappa$ reproducing, and yields the celebrated ``kernel trick'':
$\kappa(\vect{y}_1, \vect{y}_2) = \innerp{\kappa(\vect{y}_1,
  \cdot)}{\kappa(\vect{y}_2, \cdot)}_{\mathcal{H}} =
\innerp{\varphi(\vect{y}_1)}{\varphi(\vect{y}_2)}_{\mathcal{H}}$, for any
$\vect{y}_1, \vect{y}_2 \in \Real^q$.

Popular examples of reproducing kernels are:
\begin{enumerate*}

\item The linear
  $\kappa_{\text{lin}}(\vect{y}_1, \vect{y}_2) \coloneqq \vect{y}_1^{\intercal}
  \vect{y}_2$, where space $\mathcal{H}$ is nothing but $\Real^q$;

\item the Gaussian
  $\kappa_{\text{G}; \sigma}(\vect{y}_1, \vect{y}_2) \coloneqq
  \exp[-\norm{\vect{y}_1 - \vect{y}_2}^2 / (2\sigma^2)]$, where
  $\sigma\in\RealPP$ and $\norm{\cdot}$ is the standard Euclidean norm. In this
  case, $\mathcal{H}$ is infinite dimensional~\citep{Slavakis.OL.BookChapter.14};

\item the Laplacian
  $\kappa_{\text{L}; \sigma}(\vect{y}_1, \vect{y}_2) \coloneqq \exp[-\norm{\vect{y}_1 -
    \vect{y}_2}_1 / \sigma]$, where $\norm{\cdot}_1$ stands for the
  $\ell_1$-norm~\citep{Sriperumbudur.10}; and

\item the polynomial
  $\kappa_{\text{poly}; r}(\vect{y}_1, \vect{y}_2) \coloneqq
  (\vect{y}_1^{\intercal} \vect{y}_2 + 1)^r$, for some parameter
  $r\in\IntegerPP$.

\end{enumerate*}
There are several ways of generating reproducing kernels via certain operations
on well-known kernel functions such as convex combinations, products,
\etc~\citep{Scholkopf.Smola.Book}.

Define $\mathcal{H}^p$, for some $p\in \IntegerPP$, as the space whose points
take the following form:
$\bm{g} \coloneqq [g_1, \ldots, g_p]^{\intercal} \in \mathcal{H}^p$ s.t.\
$g_j\in\mathcal{H}$, $\forall j\in\overline{1,p}$, where $\overline{1,p}$ is a
compact notation for $\{1, \ldots, p\}$. For $p'\in \IntegerPP$ and given a
matrix $\vect{A}\coloneqq [a_{ij}] \in \Real^{p'\times p}$, the product
$\vect{A}\bm{g}\in \mathcal{H}^{p'}$ stands for the vector-valued function whose
$i$th entry is $\sum_{j=1}^p a_{ij}g_j$. Similarly, define
$\mathcal{H}^{p_1 \times p_2}$, for some $p_1, p_2\in \IntegerPP$, as the space
comprising all
\begin{align*}
  \bm{\mathcal{G}} \coloneqq
  \begin{bmatrix}
    g_{11} & \cdots & g_{1p_2} \\
    \vdots & \ddots & \vdots \\
    g_{p_11} & \cdots & g_{p_1p_2}
  \end{bmatrix}
  \in \mathcal{H}^{p_1 \times p_2} \,,
\end{align*}
s.t.\ $g_{ij}\in\mathcal{H}$, $\forall i\in\overline{1,p_1}$,
$\forall j\in\overline{1,p_2}$. Moreover, given
$\bm{\mathcal{G}}\in \mathcal{H}^{p_1 \times p}$ and
$\bm{\mathcal{G}}' \in \mathcal{H}^{p \times p_2}$, define the ``product''
$\bm{\mathcal{G}} \Kprod \bm{\mathcal{G}}'$ as the $p_1\times p_2$ matrix whose
$(i,j)$th entry is
$[\bm{\mathcal{G}} \Kprod \bm{\mathcal{G}}']_{ij} \coloneqq
\sum\nolimits_{l=1}^p \innerp{g_{il}}{g_{lj}'}_{\mathcal{H}}$. In the case where
$g_{il} \coloneqq \varphi(\vect{y}_{il}) = \kappa(\vect{y}_{il}, \cdot)$ and
$g_{lj}' \coloneqq \varphi(\vect{y}_{lj}') = \kappa(\vect{y}_{lj}', \cdot)$, for
some $\vect{y}_{il}, \vect{y}_{lj}'$, as in \eqref{low.rank.formula}, then the
kernel trick suggests that the previous formula simplifies to
$[\bm{\mathcal{G}} \Kprod \bm{\mathcal{G}}']_{ij} = \sum_{l=1}^p
\kappa(\vect{y}_{il}, \vect{y}_{lj}')$.

\section{Proof of Proposition~\ref{Prop:O}}\label{App:prove.prop}

By considering a probability space $(\Omega, \Sigma, \Prob)$, a basis
$\Set{e_n}_{n\in\IntegerPP}$ of $\mathcal{H}$, and by omitting most of the
entailing measure-theoretic details, the expectation of
$g = \sum_{n\in\IntegerPP} \gamma_ne_n \in \mathcal{H}$, where
$\Set{\gamma_n}_{n\in\IntegerPP}$ are real-valued RVs, is defined as
$\Expect(g) \coloneqq \sum_{n\in\IntegerPP} \Expect(\gamma_n)e_n$, provided that
the latter sum converges in $\mathcal{H}$. Conditional expectations are
similarly defined. All of the expectations appearing in this manuscript are
assumed to exist. Due to the linearity of the inner product
$\innerp{\cdot}{\cdot}_{\mathcal{H}}$, it can be verified that the conditional
expectation
$\Expect\{\innerp{g}{g'}_{\mathcal{H}} \given g'\} = \Expect\{ \sum_{n, n'}
\gamma_n \gamma_{n'} \innerp{e_n}{e_{n'}}_{\mathcal{H}} \given g' \} = \sum_{n'}
\gamma_{n'} \sum_n \Expect\{ \gamma_n \given g' \}
\innerp{e_n}{e_{n'}}_{\mathcal{H}} = \innerp{\sum_n \Expect\{ \gamma_n \given g'
  \} e_n}{ \sum_{n'} \gamma_{n'}e_{n'}}_{\mathcal{H}} = \innerp{\Expect\{g\given
  g'\}}{g'}_{\mathcal{H}}$, and
$\Expect\{\innerp{g}{g'}_{\mathcal{H}}\} =
\innerp{\Expect(g)}{\Expect(g')}_{\mathcal{H}}$ in the case where $g$ and $g'$
are independent. It can be similarly verified that these properties, which hold
for the inner product $\innerp{\cdot}{\cdot}$, are inherited by $\Kprod$.

Induction on \eqref{KARMA} suggests that $\forall \tau\in\IntegerP$,
$\bm{\varphi}_{t+\tau} = \vect{CA}^{\tau} \bm{\psi}_t + \sum_{k=1}^{\tau}
\vect{CA}^{\tau-k} \bm{\omega}_{t+k} + \bm{\upsilon}_{t+\tau}$, where
$\sum_{k=1}^{0} \vect{CA}^{-k} \bm{\omega}_{t+k} \coloneqq \vect{0}$. Then,
\begin{align}
  \bm{f}_t
  &\coloneqq \left[ \bm{\varphi}_t^{\intercal}, \bm{\varphi}_{t+1}^{\intercal},
    \ldots, \bm{\varphi}_{t+m-1}^{\intercal} \right]^{\intercal} \notag  \\&= \vect{O} \bm{\psi}_t + \bm{e}_t \,, \label{phi.O}
\end{align}
where
\begin{align*}
  \bm{e}_t \coloneqq \begin{bmatrix}
    \bm{\upsilon}_t \\
    \vect{C} \bm{\omega}_{t+1} + \bm{\upsilon}_{t+1} \\
    \sum_{k=1}^2 \vect{CA}^{2-k} \bm{\omega}_{t+k} + \bm{\upsilon}_{t+2} \\
    \vdots \\
    \sum_{k=1}^{m-1} \vect{CA}^{m-1-k} \bm{\omega}_{t+k} + \bm{\upsilon}_{t+m-1}
  \end{bmatrix} \in \mathcal{H}^{mN} \,.
\end{align*}

By observing that
$\bm{\mathcal{F}}_t = [\bm{f}_t, \bm{f}_{t+1}, \ldots, \bm{f}_{t+\tau_{\text{f}}
  -1}]$, it can be verified that
$\bm{\mathcal{F}}_t = \vect{O} \left[\bm{\psi}_t, \bm{\psi}_{t+1}, \ldots,
  \bm{\psi}_{t+\tau_{\text{f}} -1} \right] + \left[\bm{e}_t, \bm{e}_{t+1},
  \ldots, \bm{e}_{t+\tau_{\text{f}} -1} \right]$. Moreover, notice that
$\bm{\mathcal{B}}_t = [\bm{b}_t, \bm{b}_{t+1}, \ldots, \bm{b}_{t+\tau_{\text{f}}
  -1}]$, where
$\bm{b}_t \coloneqq \left[ \bm{\varphi}_t^{\intercal},
  \bm{\varphi}_{t-1}^{\intercal}, \ldots,
  \bm{\varphi}_{t-\tau_{\text{b}}+1}^{\intercal} \right]^{\intercal}
\in\mathcal{H}^{\tau_{\text{b}}N}$. Hence,
\begin{alignat*}{2}
  &&& \hspace{-30pt} \tfrac{1}{\tau_{\text{f}}}
  \bm{\mathcal{F}}_{t+1} \Kprod \bm{\mathcal{B}}_t^{\intercal} \\
  & {} = {} && \tfrac{1}{\tau_{\text{f}}} \vect{O} \left[\bm{\psi}_{t+1},
    \ldots, \bm{\psi}_{t+\tau_{\text{f}}} \right] \Kprod
  \bm{\mathcal{B}}_t^{\intercal} \\ 
  &&& + \tfrac{1}{\tau_{\text{f}}} \left[\bm{e}_{t+1}, \ldots,
    \bm{e}_{t+\tau_{\text{f}}} \right] \Kprod \bm{\mathcal{B}}_t^{\intercal} \\
  & = && \vect{O}\, \tfrac{1}{\tau_{\text{f}}} \sum\nolimits_{l=1}^{\tau_{\text{f}}}
  \bm{\psi}_{t+l} \Kprod \bm{b}_{t+l-1}^{\intercal} \\
  &&& + \tfrac{1}{\tau_{\text{f}}} \sum\nolimits_{l=1}^{\tau_{\text{f}}}
  \bm{e}_{t+l} \Kprod \bm{b}_{t+l-1}^{\intercal} \\
  & = && \vect{O}\, \tfrac{1}{\tau_{\text{f}}} \sum\nolimits_{l=1}^{\tau_{\text{f}}}
  \bm{\psi}_{t+l} \Kprod [\bm{\psi}_{t+l-1}^{\intercal} \vect{C}^{\intercal},
  \ldots, \bm{\psi}_{t+l-\tau_{\text{b}}}^{\intercal} \vect{C}^{\intercal}] \\
  &&& + \vect{O}\, \tfrac{1}{\tau_{\text{f}}} \sum\nolimits_{l=1}^{\tau_{\text{f}}}
  \bm{\psi}_{t+l} \Kprod [\bm{\upsilon}_{t+l-1}^{\intercal}, \ldots,
  \bm{\upsilon}_{t+l-\tau_{\text{b}}}^{\intercal}] \\
  &&& + \tfrac{1}{\tau_{\text{f}}} \sum\nolimits_{l=1}^{\tau_{\text{f}}}
  \bm{e}_{t+l} \Kprod \bm{b}_{t+l-1}^{\intercal} \,,
\end{alignat*}
and \eqref{low.rank.formula} is established under the following definitions:
\begin{alignat}{2}
  \bm{\Pi}_{t+1}
  & {} \coloneqq {} &&  \tfrac{1}{\tau_{\text{f}}}
  \sum\nolimits_{l=1}^{\tau_{\text{f}}}
  \bm{\psi}_{t+l} \Kprod [\bm{\psi}_{t+l-1}^{\intercal} \vect{C}^{\intercal},
  \ldots, \bm{\psi}_{t+l-\tau_{\text{b}}}^{\intercal} \vect{C}^{\intercal}] \,,
  \notag \\
  \bm{\mathcal{E}}_{t+1}^{\tau_{\text{f}}} 
  & \coloneqq && \vect{O}\, \tfrac{1}{\tau_{\text{f}}}
  \sum\nolimits_{l=1}^{\tau_{\text{f}}} \bm{\psi}_{t+l} \Kprod
  [\bm{\upsilon}_{t+l-1}^{\intercal}, \ldots, 
  \bm{\upsilon}_{t+l-\tau_{\text{b}}}^{\intercal}] \notag \\
  &&& + \tfrac{1}{\tau_{\text{f}}} \sum\nolimits_{l=1}^{\tau_{\text{f}}}
  \bm{e}_{t+l} \Kprod \bm{b}_{t+l-1}^{\intercal} \,. \label{E}
\end{alignat}

By virtue of the independency between $(\bm{\psi}_t)_t$ and
$(\bm{\upsilon}_t)_t$, the zero-mean assumption on $(\bm{\upsilon}_t)_t$, as
well as standard properties of the conditional
expectation~\citep[\S9.7(k)]{Williams.book} with respect to independency, it can
be verified that
\begin{align}
  & \Expect\{ \bm{\psi}_{t+l} \Kprod
  [\bm{\upsilon}_{t+l-1}^{\intercal}, \ldots, 
  \bm{\upsilon}_{t+l-\tau_{\text{b}}}^{\intercal}] \given \bm{\psi}_{t+l} \}
    \notag \\
  & = \bm{\psi}_{t+l} \Kprod [ \Expect\{\bm{\upsilon}_{t+l-1}^{\intercal}\}, \ldots, 
  \Expect\{\bm{\upsilon}_{t+l-\tau_{\text{b}}}^{\intercal}\}] = \vect{0}
    \,. \label{E1}
\end{align}
Moreover, for any $i\in\overline{1,m}$ and any
$j\in\overline{1,\tau_{\text{b}}}$, the $(i,j)$th $N\times N$ block of the
second term in the expression of $\bm{\mathcal{E}}_{t+1}^{\tau_{\text{f}}}$ in
\eqref{E} becomes equal to
\begin{align}
  & \sum\nolimits_{k=1}^{i-1} \vect{CA}^{i-1-k}
    \tfrac{1}{\tau_{\text{f}}} \sum\nolimits_{l=1}^{\tau_{\text{f}}}
    \bm{\omega}_{t+l+k} \Kprod \bm{\varphi}_{t+l-j}^{\intercal} \notag \\
  & \hphantom{=\ } + \tfrac{1}{\tau_{\text{f}}} \sum\nolimits_{l=1}^{\tau_{\text{f}}}
    \bm{\upsilon}_{t+l+i-1} \Kprod \bm{\varphi}_{t+l-j}^{\intercal} \notag \\
  & = \sum\nolimits_{k=1}^{i-1} \vect{CA}^{i-1-k}
    \tfrac{1}{\tau_{\text{f}}} \sum\nolimits_{l=1}^{\tau_{\text{f}}}
    \bm{\omega}_{t+l+k} \Kprod \bm{\psi}_{t+l-j}^{\intercal}
    \vect{C}^{\intercal} \notag \\
  & \hphantom{=\ } + \sum\nolimits_{k=1}^{i-1} \vect{CA}^{i-1-k}
    \tfrac{1}{\tau_{\text{f}}} \sum\nolimits_{l=1}^{\tau_{\text{f}}}
    \bm{\omega}_{t+l+k} \Kprod \bm{\upsilon}_{t+l-j}^{\intercal} \notag \\  
  & \hphantom{=\ } + \tfrac{1}{\tau_{\text{f}}}
    \sum\nolimits_{l=1}^{\tau_{\text{f}}} \bm{\upsilon}_{t+l+i-1} \Kprod
    \bm{\psi}_{t+l-j}^{\intercal} \vect{C}^{\intercal} \notag \\
  & \hphantom{=\ } + \tfrac{1}{\tau_{\text{f}}}
    \sum\nolimits_{l=1}^{\tau_{\text{f}}} \bm{\upsilon}_{t+l+i-1} \Kprod
    \bm{\upsilon}_{t+l-j}^{\intercal}  \,. \label{block.E2}
\end{align}
Since $t+l+k > t+l > t+l-j$ and $t+l+i-1 \geq t+l > t+l-j$, $\bm{\psi}_{t+l-j}$
precedes $\bm{\omega}_{t+l+k}$ on the time axis, while $\bm{\upsilon}_{t+l+i-1}$
precedes $\bm{\upsilon}_{t+l-j}$. Hence, due to independency,
$\Expect\{ \bm{\omega}_{t+l+k} \Kprod \bm{\psi}_{t+l-j}^{\intercal} \given
\bm{\psi}_{t'} \} = \Expect\{ \bm{\omega}_{t+l+k} \given \bm{\psi}_{t'} \}
\Kprod \bm{\psi}_{t+l-j}^{\intercal} = \Expect\{ \bm{\omega}_{t+l+k} \} \Kprod
\bm{\psi}_{t+l-j}^{\intercal} = \vect{0}$, and
$\Expect \{ \bm{\upsilon}_{t+l+i-1} \Kprod \bm{\upsilon}_{t+l-j}^{\intercal}
\given \bm{\psi}_{t'} \} = \Expect \{ \bm{\upsilon}_{t+l+i-1}\} \Kprod \Expect\{
\bm{\upsilon}_{t+l-j}^{\intercal} \} = \vect{0}$. It can be also similarly
verified that
$\Expect\{ \bm{\omega}_{t+l+k} \Kprod \bm{\upsilon}_{t+l-j}^{\intercal} \given
\bm{\psi}_{t'} \} = \vect{0}$ and
$\Expect\{ \bm{\upsilon}_{t+l+i-1} \Kprod \bm{\psi}_{t+l-j}^{\intercal} \given
\bm{\psi}_{t'} \} = \vect{0}$. As a result, the conditional expectation of
\eqref{block.E2}, given $\bm{\psi}_{t'}$, becomes $\vect{0}$. This observation
and \eqref{E1} establish claim \eqref{cond.expectation} of the proposition.

Under the assumptions on wide-sense stationarity, the covariance sequences of
the processes $(\bm{\omega}_t \Kprod \bm{\psi}_{t-\tau}^{\intercal})_t$,
$(\bm{\omega}_t \Kprod \bm{\upsilon}_{t-\tau}^{\intercal})_t$,
$(\bm{\upsilon}_t \Kprod \bm{\psi}_{t-\tau}^{\intercal})_t$,
$( \bm{\psi}_t \Kprod \bm{\upsilon}_{t-\tau}^{\intercal})_t$,
$(\bm{\upsilon}_t \Kprod \bm{\upsilon}_{t-\tau}^{\intercal})_t$,
$\forall\tau \in\IntegerPP$, are summable over all lags; in fact, the
covariances of non-zero lags become zero due to the assumptions on
independency. Hence, by the mean-square ergodic theorem~\citep{Petersen}, sample
averages of the previous processes converge in the mean-square
($\mathcal{L}_2$-) sense to their ensemble means. For example, applying
$\lim_{\tau_{\text{f}} \to\infty}$, in the mean-square sense, to the first part
of $\bm{\mathcal{E}}_{t+1}^{\tau_{\text{f}}}$ in \eqref{E} and by
recalling standard properties of the conditional
expectation~\citep[\S9.7(a)]{Williams.book} yield
\begin{align}
  & \vect{O} \lim_{\tau_{\text{f}} \to\infty} \tfrac{1}{\tau_{\text{f}}}
  \sum\nolimits_{l=1}^{\tau_{\text{f}}} \bm{\psi}_{t+l} \Kprod
  [\bm{\upsilon}_{t+l-1}^{\intercal}, \ldots, 
    \bm{\upsilon}_{t+l-\tau_{\text{b}}}^{\intercal}] \notag \\
  & = \vect{O} \Expect\{ \bm{\psi}_{t+l} \Kprod
    [\bm{\upsilon}_{t+l-1}^{\intercal}, \ldots, 
    \bm{\upsilon}_{t+l-\tau_{\text{b}}}^{\intercal}] \} \notag \\
  & = \vect{O} \Expect\{ \Expect\{ \bm{\psi}_{t+l} \Kprod
  [\bm{\upsilon}_{t+l-1}^{\intercal}, \ldots, 
  \bm{\upsilon}_{t+l-\tau_{\text{b}}}^{\intercal}] \given \bm{\psi}_{t+l} \} \}
    = \vect{0} \,. \label{E1.ergodicity}
\end{align}
By following similar arguments, it can be verified that the application of
$\lim_{\tau_{\text{f}} \to\infty}$ to \eqref{block.E2} renders the second part
of \eqref{E} equal to $\vect{0}$. This finding and \eqref{E1.ergodicity}
establish the final claim of the proposition.

\bibliographystyle{elsarticle-num-names}
\bibliography{library.bib}


\end{document}